\documentclass[review]{elsarticle}

\usepackage{lineno,hyperref}
\modulolinenumbers[5]
\usepackage{url}
\usepackage{amssymb,endnotes,setspace,enumerate,color,xspace,lscape}
\usepackage{amsmath}
\usepackage{graphicx}
\usepackage{amsthm,subfigure,graphics,epsfig}
\usepackage{bbm}
\usepackage{dsfont}
\usepackage{multirow}
\usepackage{epstopdf}
\usepackage{float}
\usepackage{hyperref, bm}
\usepackage[symbol]{footmisc}
\usepackage[table]{xcolor}
\usepackage{xr}
\usepackage{caption}
\usepackage{appendix}
\usepackage{afterpage}
\usepackage{float}
\usepackage{accents}
\usepackage[ruled,vlined]{algorithm2e}

\DeclareMathOperator*{\argmax}{arg\,max}
\DeclareMathOperator*{\argmin}{arg\,min}
\journal{Neurocomputing}

\begin{document}
\nolinenumbers
\begin{frontmatter}

\title{Deep Feature Screening: Feature Selection for Ultra High-Dimensional Data via Deep Neural Networks}

\author{Kexuan Li\corref{mycorrespondingauthor}}
\address{Global Analytics and Data Sciences, Biogen}
\cortext[mycorrespondingauthor]{Corresponding author}
\ead{kexuan.li@biogen.com}
\author{Fangfang Wang}
\ead{fwang4@wpi.edu}
\author{Lingli Yang}
\ead{lyang8@wpi.edu}
\address{Department of Mathematical Sciences, Worcester Polytechnic Institute}
\author{Ruiqi Liu}
\ead{ruiqliu@ttu.edu}
\address{Department of Mathematics and Statistics, Texas Tech University}

\begin{abstract}
The applications of traditional statistical feature selection methods to high-dimension, low-sample-size data often struggle and encounter  challenging problems, such as overfitting, curse of dimensionality, computational infeasibility, and strong model assumptions. In this paper, we propose a novel two-step nonparametric approach called Deep Feature Screening (DeepFS) that can overcome these problems and  identify significant features with high precision for ultra high-dimensional, low-sample-size data. This approach first extracts a low-dimensional representation of input data and then applies feature screening on the original input feature space based on multivariate rank distance correlation recently developed by Deb and Sen \cite{multi_rank_JASA}.
This approach combines the strengths of both deep neural networks and feature screening, thereby  having the following appealing features in addition to its ability of handling ultra high-dimensional data with small number of samples: (1) it is model free and distribution free; (2) it can be used for both supervised and unsupervised feature selection; and (3) it is capable of recovering the original input data.  The superiority of DeepFS is demonstrated via extensive simulation studies and real data analyses.
\end{abstract}


\end{frontmatter}


\section{Introduction}
(Ultra) high-dimensional data are  commonly encountered in research areas such as machine learning, computer version, financial engineering, and biological science. In the literatures of statistics and machine learning, high dimension is referred to the case where the dimension (or the number of features)  $p$ grows to infinity, while the ultra-high dimension means that the dimension $p$ grows at a non-polynomial rate of sample size $n$ (say, $p = O(\exp(n^\xi))$ for some $\xi>0$; see \cite{Fan_1} and \cite{Fan_2}). To analyze (ultra) high-dimensional data, feature selection has been regarded as a powerful tool to achieve dimension reduction, in that a significant amount of  features are irrelevant and redundant  in many problems of interest. Correctly selecting a representative subset of features plays an important role in these applications. For example, in genome-wide association studies (GWAS), researchers are interested in identifying genes that contain single-nucleotide polymorphisms (SNPs) associated with certain target human diseases. GWAS datasets oftentimes contain a large number of SNPs (e.g., $p\geq10^5$), but the sample size $n$ is relatively small (e.g., $n\leq10^3$). It is well known that the process of selecting an optimal subset is an NP-hard problem \cite{NP}. Thus, a vast amount of feature selection methods have been proposed to address this problem.

Based on the interaction of feature selection search and the learning model, the traditional feature selection methods can be broadly categorized into three classes: filter method, wrapper method, and embedded method. The filter method. requires features to be selected according to a certain statistical measure such as information gain, chi-square test, fisher score, correlation coefficient, or variance threshold. In the wrapper method, features are selected based on a classifier; some commonly adopted  classifiers include recursive feature elimination and sequential feature selection algorithms. The embedded method  uses ensemble learning and hybrid learning methods for feature selection. See \cite{feasure_selection_survey_4}, \cite{kabir2011new}, \cite{feasure_selection_survey_1}, \cite{feasure_selection_survey_3},   \cite{feasure_selection_survey_2},  \cite{mohsenzadeh2016incremental},  \cite{cilia2019variable}, \cite{barbiero2021predictable} and  \cite{mirzaei2017variational} for more detail on the three methods. However, these classical methods suffer from some potential problems when applied to ultra high-dimensional data. For example,  they ignore feature dependence and nonlinear structure, lack flexibility, and require a large sample size. When implementing these methods, people tend to adopt  a parametric form (like linear model and logistic regression model) to describe the relationship between  response and  features,  ignoring the interactions among features, which, however, can be complex and nonlinear in practice \cite{LassoWu, pLasso}. Despite that both supervised and unsupervised feature selections have important applications, most algorithms cannot combine  supervised and unsupervised learnings effectively
(see \cite{feasure_selection_survey_unsurpervised_3, feasure_selection_survey_unsurpervised_2, neucomputing_unsupervised, neucomputing_zhu2018co, neucomputing_Boltzmann, feasure_selection_survey_unsurpervised_1}). The successful application of feature selection algorithms to high-dimensional data relies on a large sample size. However, when dealing with high-dimension, low-sample-size data, such methods suffer from computational instability in addition to curse of dimensionality and overfitting.

To overcome these problems, we propose an effective feature selection approach based on deep neural network and feature screening, called DeepFS, under an ultra-high-dimension, small-sample-size setup with only a few tuning parameters. Our method enjoys  several advantages: (1) it can be used for both supervised and unsupervised feature selection; (2) it is distribution-free and model-free; (3) it can capture nonlinear and interaction among features; (4) it can provide an estimation of the number of active features. To be more specific, our feature selection procedure consists of two steps: feature extraction and feature screening. In the first step, we use an autoencoder or supervised autoencoder to extract a good representation of data. In the second step, we screen each feature and compute their corresponding importance scores associated with the low-dimensional representation generated from the first step by means of multivariate rank distance correlation. The features with high importance scores will then be selected. Apparently, this two-step method has the advantages of both deep learning and feature screening. It is worth emphasizing that unlike other feature selection methods which focus on selecting features from the extracted feature space, our feature selection process is applied to the original input feature space. To the best of our knowledge, this study is the first attempt to combine deep neural networks and feature screening.

The rest of the paper is organized as follows. In Section \ref{literiture_review}, we formulate the problem of interest and briefly review the feature screening and feature selection in deep learning literature. We introduce our method in Section \ref{Method} and comprehensive simulation studies are presented in Section \ref{Simulation}.
In Section \ref{Real_Data}, we apply DeepFS to some real-world datasets and give our conclusion in Section \ref{Conclusion}.

\section{Problem Formulation and Related Works} \label{literiture_review}
In this section, we formulate the problem under study, followed by a literature review of feature selection methods by means of feature screening and deep learning.

\subsection{Problem Formulation}
We first describe the problem often encountered in unsupervised feature selection and supervised feature selection.
Suppose there are $n$ observations $\boldsymbol{x}_i \in \mathbb{R}^p, i=1, \ldots, n$, that are independent and identically distributed ($i.i.d.$) from a distribution $p(\boldsymbol{x})$, $\boldsymbol{x} \in  \mathbb{R}^p$. In unsupervised feature selection, the goal is to identify a subset $\mathcal{S}  \subseteq \{1, 2, \ldots, p\}$ of the most discriminative and informative features with size $|\mathcal{S}| = k \leqslant p$ and also a reconstruction function $g: \mathbb{R}^k \rightarrow \mathbb{R}^p$ from a low dimensional feature space $\mathbb{R}^k$ to the original feature space $\mathbb{R}^p$,
such that the expected loss between $g(\boldsymbol{x}^{(\mathcal{S})}) \in \mathbb{R}^p$ and the original input $\boldsymbol{x} \in  \mathbb{R}^p$  is minimized, where $\boldsymbol{x}^{(\mathcal{S})} = (x_{s_1}, \ldots, x_{s_k})^{\top} \in \mathbb{R}^k$ and $s_i \in \mathcal{S}$. 
In supervised feature selection, people not only know the sample design matrix $(\boldsymbol{x}_1, ...,\boldsymbol{x}_n)^{\top} \in \mathbb{R}^{n \times p}$, but also the label vector $\boldsymbol{y} = (y_1,...,y_n)^{\top} \in \mathbb{R}^n$, where $y_i$ can be  continuous or categorical.  Suppose that the true relationship between a subset of features, $\mathcal{S}$, and the label is given by $y_i = f(\boldsymbol{x}_i^{(\mathcal{S})})$, with $\boldsymbol{x}_i^{(\mathcal{S})} = (x_{i, s_1}, \ldots, x_{i, s_k})^{\top}, s_i \in \mathcal{S}$, $i=1, \ldots, n$.  For example, we can assume $f(\cdot)$ has a parametric linear form: $y_i = f(\boldsymbol{x}_i^{(\mathcal{S})}) = \beta_0 + \sum_{j=1}^k\beta_j x_{i, s_j} + \epsilon_i$ for a continuous response or logistic regression $\log \frac{Pr(y_i = 1)}{Pr(y_i = 0)} = \beta_0 + \sum_{j=1}^k\beta_j x_{i, s_j}$ for a categorical response.  To better capture the complex relationship between the response and the features in real world applications, we would consider more flexible structures for $f(\cdot)$, such as neural networks.

\subsection{Related Works}

The literature on feature selection is vast and encompasses many fields. In this subsection, we shall not provide a comprehensive review, but rather focus on popular methods in feature screening and deep learning. We refer the reader to \cite{kumar2014feature} for a more in-depth review of the classical feature selection literature.

Feature screening is a fundamental problem in the analysis of ultra high-dimensional data. To be  specific, feature screening is a process of assigning a numerical value, known as importance score, to each feature according to a certain statistical measure that quantifies the strength of dependence between the feature and the response, and then ranking the importance scores and  selecting the top $k$ features accordingly. The statistical measure can be Pearson correlation \cite{Fan_1} and distance correlation \cite{Distance_correlation_Runzeli}, among others. A good feature screening method should enjoy the theoretical property called \emph{sure screening property},  meaning that all the true features can be selected with probability approaching  one as the sample size goes to infinity. It is  worth mentioning that \cite{multi_rank_JASA} proposes a novel test statistic called multivariate rank distance correlation that makes it possible to carry out a model-free nonparamatric procedure to test mutual independence between random vectors in multi-dimensions.  The authors in \cite{zhao2021distribution} apply the test statistic of \cite{multi_rank_JASA} to feature screening and introduce a distribution-free nonparametric  screening approach called MrDc-SIS that is proved to be asymptotic sure screening consistency. However, similar to other feature screening methods, MrDc-SIS has several drawbacks: (1) It ignores the reconstruction of  original inputs; (2) It does not consider inter-dependence among the features; (3) It is not trivial to  extend MrDc-SIS to unsupervised feature selection or supervised feature selection with categorical responses. Thus, MrDc-SIS is of limited effectiveness.

In recent years, deep learning has made a great breakthrough in both theory and practice.  In particular, it has been proven that deep neural networks can achieve the minimax rate of convergence in a nonparametric setting under some mild conditions \cite{DIVE, schmidt2020nonparametric, liu2022optimal}.  The work of \cite{Literature_Review_Liang} studies the rate of convergence for deep feedforward neural nets in semiparametric inference.
The successful applications include, but are not limited to, computer vision \cite{Literature_Review_CV}, natural language processing \cite{Literature_Review_NLP}, spatial statistics \cite{li2023semiparametric}, survival analysis \cite{li2022variable}, drug discovery and toxicology \cite{Literature_Review_drug}, and dynamics system \cite{ODE}, functional data analysis \cite{wang2021estimation}.

Applying deep learning to feature selection has also gained much attention. For example, deep feature selection (DFS) of \cite{DFS} learns one-to-one connections between input features and the first hidden layer nodes. Using a similar idea, \cite{DNP} proposes a so-called deep neural pursuit (DNP) that selects relevant features by averaging out gradients with lower variance via multiple dropouts. However, both DFS and DNP ignore the reconstruction of original input and use only a simple multi-layer perceptron, which fail to capture the complex structure.

Another technique commonly used in deep learning for the purpose of feature selection is reparametrization trick, or Gumbel-softmax trick \cite{gumbel_softmax}. For example, \cite{CAE} and \cite{FsNet} consider a concrete selector layer such that the gradients can pass through the network for discrete feature selection. By using reparametrization trick, the network can stochastically select a linear combination of inputs, which leads to a discrete subset of features in the end.  However, caution is called for when one uses a concrete random variable for discrete feature selection.  First, the performance is very sensitive to the tuning parameters. Second, because the concrete selector layer stochastically searches $k$ out of $p$ features, where $k$ is predetermined representing the number of features one desires to select, the probability of concrete selector layer picking up correct features is very small when $p$ is large relative to $k$.  Third, successfully training concrete selector layer requires a huge amount of samples, which is  infeasible under high-dimension, low-sample-size setting.

Recently, \cite{lassonet} introduces a new feature selection framework for neural networks called LassoNet by adding a residual layer between the input layer and the output, penalizing the parameters in the residual layer, and imposing a constraint that the norm of the parameters in the first layer is less than the corresponding norm of the parameters in the residual layer.  The authors have demonstrated that LassoNet ``significantly outperforms state-of-the-art methods for feature selection and regression".  In this paper, we will show through both simulations and real data analyses that our proposed method performs even better than LassoNet, especially when the sample size is small (see Sections \ref{Simulation} and \ref{Real_Data}).

The study by \cite{teacher_student} considers a teacher-student scheme for feature selection. In the teacher step, a sophisticated network architecture is used in order to capture the complex hidden structure of the data. In the student step, the authors use a single-layer feed-forward neural network with a row-sparse regularization to mimic the low-dimensional data generated from the teacher step and then perform feature selection. The idea of adding a row-sparse regularization to hidden layers in feature selection is not new (e.g., \cite{neucomputing_group, autoencoder_feature_selection, feng2018graph}). However, when dealing with high-dimension, low-sample-size data, these techniques fail to select correct features because of overfitting and high-variance gradients. Moreover, their performance is very sensitive to the regularization parameters.

\section{Proposed Two-step Feature Selection Framework} \label{Method}

In this section, we provide the details of our method.  It consists of two step.  In Step 1, we use an autoencoder to extract a low-dimensional representation of the original data and this is known as feature extraction step, while in Step 2, we apply feature screening via multivariate rank distance correlation learning to achieve  feature selection.

\subsection{Step 1: Dimension Reduction and Feature Extraction}
In the first step, we use an autoencoder to learn a complex representation of the input data. Autoencoder is one type of feed-forward neural networks, and is commonly used for dimension reduction.  A standard (unsupervised) autoencoder consists of two parts, the encoder and the decoder. Suppose both the input space and output space is $\mathcal{X}$ and the hidden layer space is $\mathcal{F}$.  Throughout the paper, we assume $\mathcal{X} = \mathbb{R}^p$ and $\mathcal{F}=\mathbb{R}^h$. The goal is to find two maps $\Phi:\mathcal{X}\rightarrow \mathcal{F}$ and $ \Psi:\mathcal{F}\rightarrow \mathcal{X}$ that minimize the reconstruction loss function $\mathcal{L}_r(\Theta|\boldsymbol{x}) = n^{-1}\sum_{i=1}^n||\boldsymbol{x}_i - \Psi(\Phi(\boldsymbol{x}_i))||_2^2 $, where $\Theta = [\Theta_\Phi, \Theta_\Psi]$ collects all the model parameters and $||\cdot||_2$ is the $l_2$ norm. Here, we refer to $\Phi$ as encoder and $\Psi$ as decoder. To better learn the possibly nonlinear structure of the features, we assume that $\Phi$ and $\Psi$ are neural networks. For example, suppose that there is only one layer in both encoder and decoder; then for $\boldsymbol{x}\in \mathcal{X}$, $\Phi(\boldsymbol{x})=\sigma(\boldsymbol{W}\boldsymbol{x}+b)\in \mathcal{F}$,  $\Psi(\Phi(\boldsymbol{x}))=\sigma'(\boldsymbol{W}'\Phi(\boldsymbol{x})+b')$, and
\begin{equation} \label{Eq Reconstruction loss}
\mathcal{L}_r(\Theta|\boldsymbol{x}) = \frac{1}{n}\sum_{i=1}^n||\boldsymbol{x}_i - \Psi(\Phi(\boldsymbol{x}_i))||_2^2 = \frac{1}{n}\sum_{i=1}^n||\boldsymbol{x}_i - \sigma'(\boldsymbol{W}'\sigma(\boldsymbol{W}\boldsymbol{x}_i+b)+b')||_2^2,
\end{equation}
where $\sigma(), \sigma'()$ are nonlinear active functions, $\boldsymbol{W}, \boldsymbol{W}'$ are weight matrices, and $b, b'$ are bias vectors. The standard autoencoder can be replaced with its variants, such as sparse autoencoder, denoising autoencoder, and variational autoencoder.

When it comes to supervised feature selection, we  use a supervised autoencoder instead of the standard (unsupervised) autoencoder. Therefore, we add an additional loss on the hidden layer, such as the mean square loss for continuous response or the cross-entropy loss for categorical response. Let $\mathcal{L}_s(\cdot)$ be the supervised loss on the hidden layer and $\mathcal{L}_r(\cdot)$ be the reconstruction loss as in Equation (\ref{Eq Reconstruction loss}). The loss for supervised autoencoder with continuous response is:
\begin{equation} \label{Eq Continuous loss}
\begin{split}
\mathcal{L}(\Theta|\boldsymbol{x}, y) &= \mathcal{L}_s(\Theta_\Phi, \Theta_\Upsilon|\boldsymbol{x}, y) + \lambda\mathcal{L}_r(\Theta_\Phi, \Theta_\Psi|\boldsymbol{x})\\
&= \frac{1}{n}\sum_{i=1}^n\left(||y_i -  \Upsilon(\Phi(\boldsymbol{x}_i))||_2^2 + \lambda||\boldsymbol{x}_i - \Psi(\Phi(\boldsymbol{x}_i))||_2^2 \right),
\end{split}
\end{equation}
where $y \in \mathcal{Y}, \Theta = [\Theta_\Phi, \Theta_\Psi, \Theta_\Upsilon]$ is the model parameters, $\Upsilon: \mathcal{F}\rightarrow \mathcal{Y}$ is the regressor, and $\lambda$ is the tuning parameter controlling the magnitude of the reconstruction loss.  It is worth mentioning that while in classical feature selection method $\mathcal{Y} = \mathbb{R}$, in our setting, $\mathcal{Y}$ could be a more general space, like $\mathbb{R}^k$ for some integer $k$ greater than 1. The corresponding loss in the supervised autoencoder with categorical response becomes:
\begin{equation} \label{Eq Coategorical loss}
\begin{split}
\mathcal{L}(\Theta|\boldsymbol{x}, y) &= \mathcal{L}_s(\Theta_\Phi, \Theta_\Upsilon|\boldsymbol{x}, y) + \lambda\mathcal{L}_r(\Theta_\Phi, \Theta_\Psi|\boldsymbol{x})\\
&= \frac{1}{n}\sum_{i=1}^n\left(-\log\left(\frac{\exp(\Upsilon(\boldsymbol{x}_i)_{y_i})}{\sum_{c=1}^C\exp(\Upsilon(\boldsymbol{x}_i)_c)}\right) + \lambda||\boldsymbol{x}_i - \Psi(\Phi(\boldsymbol{x}_i))||_2^2 \right),
\end{split}
\end{equation}
where $\Theta, \Phi$, $\Psi$ and $\lambda$ are the same as Equation (\ref{Eq Continuous loss}), $C$ is the number of classes, and $\Upsilon: \mathcal{F} \rightarrow \mathbb{R}^C$ is a classifier on the hidden layer with softmax output.

The training process in the first step is an optimization problem, i.e., minimizing $\mathcal{L}(\Theta | \boldsymbol{x}, y)$.  Once the model is trained, we can extract a subset of features by mapping the original inputs from $\mathcal{X}$ to the low dimensional hidden space $\mathcal{F}$. Precisely speaking,  we  define the normalized encoded input
$$\boldsymbol{x}_{\textrm{encode}} = \frac{\Phi(\boldsymbol{x}) - \min_{\boldsymbol{x}\in \mathcal{X}}\Phi(\boldsymbol{x})}{\max_{\boldsymbol{x}\in \mathcal{X}}\Phi(\boldsymbol{x}) - \min_{\boldsymbol{x}\in \mathcal{X}} \Phi(\boldsymbol{x})} \in \mathcal{F},$$ which generates the abstract features from the original high dimensional data and will be used later in the second step.

\subsection{Step 2:  Feature Screening}
With the aid of the low-dimensional representation of the high-dimensional features generated from the first step, in the second step, we screen all the features via multivariate rank distance correlation learning to select the relevant ones. Unlike \cite{teacher_student} which uses a single-layer feed forward network with a row-sparse regularization to mimic the data generated from the first step, we screen each feature  with respect to $\boldsymbol{x}_{\textrm{encoded}}$.  The reason for using feature screening instead of neural network is the small amount of training samples.   When  training samples are limited, neural network can easily result in overfitting; however, if we only consider one feature at a time, the dependence between the feature and $\boldsymbol{x}_{\textrm{encoded}}$ can still be well estimated even when the sample size is small.

The feature screening procedure via multivariate rank distance correlation has been proved to be of asymptotic sure screening consistency by  \cite{zhao2021distribution}. For the sake of  convenience, in what follows, we recapitulate the multivariate rank distance correlation of \cite{multi_rank_JASA}.

\begin{figure}[ht!]
  \centering
  \includegraphics[width=5in]{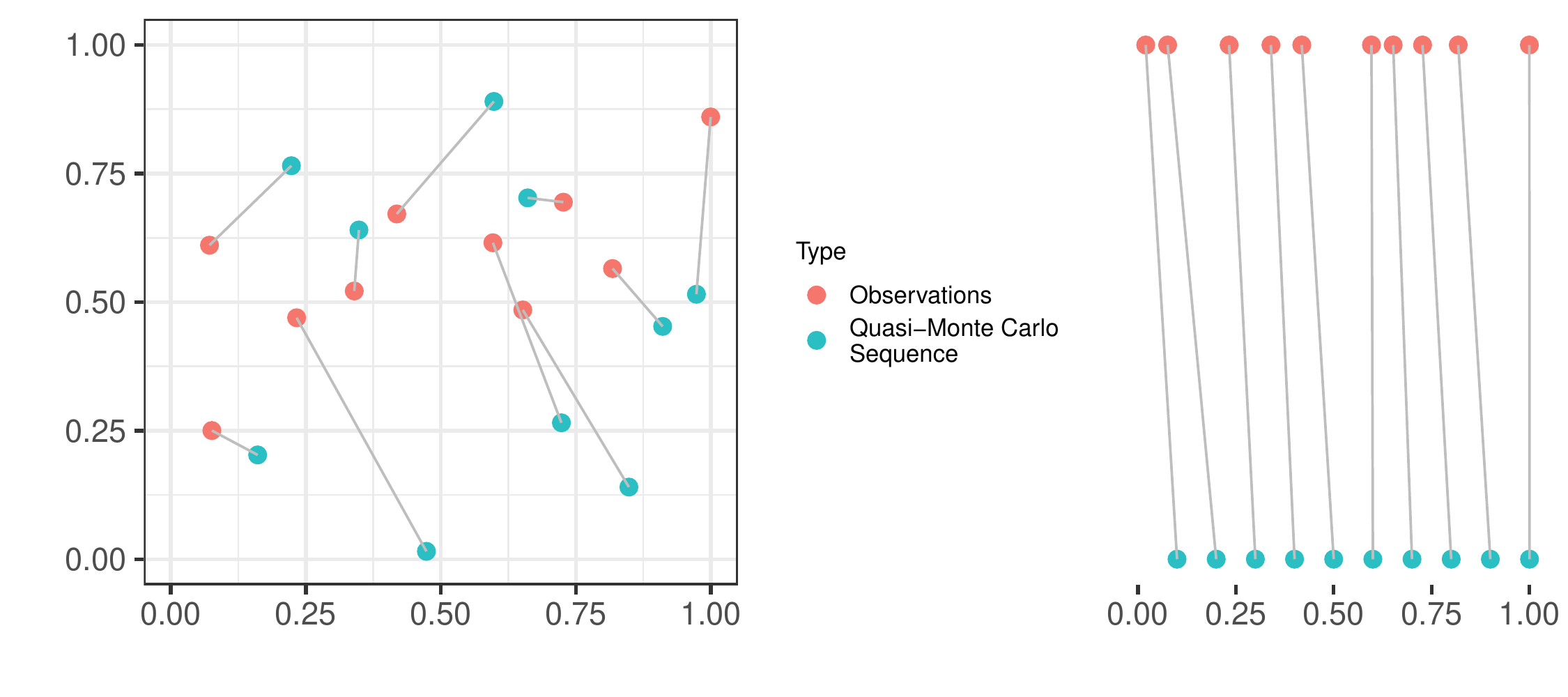}
  \caption{\it An illustration of multivariate rank.  The red dots are the observed data points and blue dots are the ``uniform like'' quasi-Monte Carlo sequence on $[0,1]^2$. The (empirical) multivariate rank of each observation is the closest point in the quasi-Monte Carlo sequence.}
  \label{figure:multivariate_rank}
\end{figure}

We start by introducing the multivariate rank.  Let $\{\mathbf{c}_1,...,\mathbf{c}_n\}\subset[0, 1]^p$ be a ``uniform-like" quasi-Monte Carlo sequence on $[0, 1]^p$ such as Halton sequences \cite{Halton}, Sobol sequence \cite{Sobol}, and  equally-spaced $p$-dimensional lattice.
Given  $i.i.d.$ random vectors $\mathbf{X}_1,...,\mathbf{X}_n \in \mathbb{R}^p$,  consider the following optimization problem:
\begin{equation}\label{assign}
      \widehat \sigma = \argmin_{\sigma=(\sigma(1),...,\sigma(n))\in S_n}\sum_{i=1}^n\parallel \mathbf{X}_i - \mathbf{c}_{\sigma(i)}\parallel^2  = \argmax_{\sigma=(\sigma(1),...,\sigma(n))\in S_n}\sum_{i=1}^n \langle\mathbf{X}_i,  \mathbf{c}_{\sigma(i)}\rangle,
\end{equation}
where $\parallel\cdot\parallel$ and $\langle \cdot, \cdot \rangle$ denote the usual Euclidean norm and inner product, and $S_n$ is the set of all permutations of $\{1, 2,..., n\}$. The (empirical) multivariate rank of $\mathbf{X}_i$ is a $p$-dimensional vector defined as
\begin{align}\label{rank}
\textrm{Rank}(\mathbf{X}_i) = \mathbf{c}_{\widehat \sigma(i)} ,\,\, \textrm{ for $i=1,\ldots,n$.}
\end{align}
Figure \ref{figure:multivariate_rank} illustrates the idea of multivariate rank on  $[0, 1]^2$.

The rank distance covariance is defined as a measure of mutual dependence. Suppose that $\{\mathbf{X}_1,\ldots ,\mathbf{X}_n\}$ and  $\{\mathbf{Y}_1, \ldots, \mathbf{Y}_n\}$ are samples of $n$ observations from $p$-dimensional random vector $\mathbf{X}$ and  $q$-dimensional random vector $\mathbf{Y}$, respectively. The (empirical) rank distance covariance and rank distance correlation between $\mathbf{X}$ and $\mathbf{Y}$ are defined as
\begin{align} \label{RdCorr}
\textrm{RdCov}^2(\mathbf{X}, \mathbf{Y}) = S_1 + S_2 -2S_3, \,\,\, ~ \textrm{RdCorr}(\mathbf{X}, \mathbf{Y}) = \frac{\textrm{RdCov}(\mathbf{X}, \mathbf{Y})}{\sqrt{\textrm{RdCov}(\mathbf{X}, \mathbf{X})\textrm{RdCov}(\mathbf{Y}, \mathbf{Y})}}
\end{align}
respectively, where
\begin{align*}
&S_1 = \frac{1}{n^2} \sum_{k,l=1}^n\parallel\textrm{Rank}(\mathbf{X}_k) - \textrm{Rank}(\mathbf{X}_l)\parallel*\parallel\textrm{Rank}(\mathbf{Y}_k) - \textrm{Rank}(\mathbf{Y}_l)\parallel,\\
&S_2 = \Big( \frac{1}{n^2} \sum_{k,l=1}^n\parallel\textrm{Rank}(\mathbf{X}_k) - \textrm{Rank}(\mathbf{X}_l)\parallel \Big)*\Big( \frac{1}{n^2} \sum_{k,l=1}^n\parallel\textrm{Rank}(\mathbf{Y}_k) - \textrm{Rank}(\mathbf{Y}_l)\parallel\Big),\\
&S_3 = \frac{1}{n^3} \sum_{k,l,m=1}^n\parallel\textrm{Rank}(\mathbf{X}_k) - \textrm{Rank}(\mathbf{X}_l)\parallel*\parallel\textrm{Rank}(\mathbf{Y}_k) - \textrm{Rank}(\mathbf{Y}_m)\parallel.
\end{align*}

Let $\boldsymbol{x}_{\textrm{encode}} = (x_{\textrm{encode}, 1}, \ldots, x_{\textrm{encode},h})^{\top}$ be the $h$-dimensional encoded data generated from the first step and $\mathbf{X} = (X_1, \ldots, X_p)^{\top} \in \mathbb{R}^p$ be the vector of original high-dimensional input features. We compute the rank distance correlation between $\boldsymbol{x}_{\textrm{encode}}$ and each feature $X_i$ as $\omega_i = \textrm{RdCorr}(X_i, \boldsymbol{x}_{\textrm{encode}})$ for $ i=1,\ldots, p$, which measures the strength of dependence between $X_i$ and  $\boldsymbol{x}_{\textrm{encode}}$. Because $\boldsymbol{x}_{\textrm{encode}}$ extracts most of the information in the original features, $\omega_i$ can also be viewed as a marginal utility that reflects the importance of the corresponding feature. In other words, the finally selected $k$ active features  are the ones whose rank distance correlations  are among the top $k$.

It is worthwhile to mention that this design also provides an estimate of the number of true active features. Let $\omega_{(1)} \leq \omega_{(2)} \leq \ldots \leq \omega_{(p)}$ be the order statistics of $\omega$s. Suppose that $\mathcal{M}$ is the set of true active features with size $m_0 \leq p$.  Apparently, $\widehat{m} = \argmax_{1\leq i\leq p-1} \omega_{(i+1)}/\omega_{(i)}$ is an estimation of $m_0$.

\begin{figure}[ht!]
  \centering
  \includegraphics[width=5in]{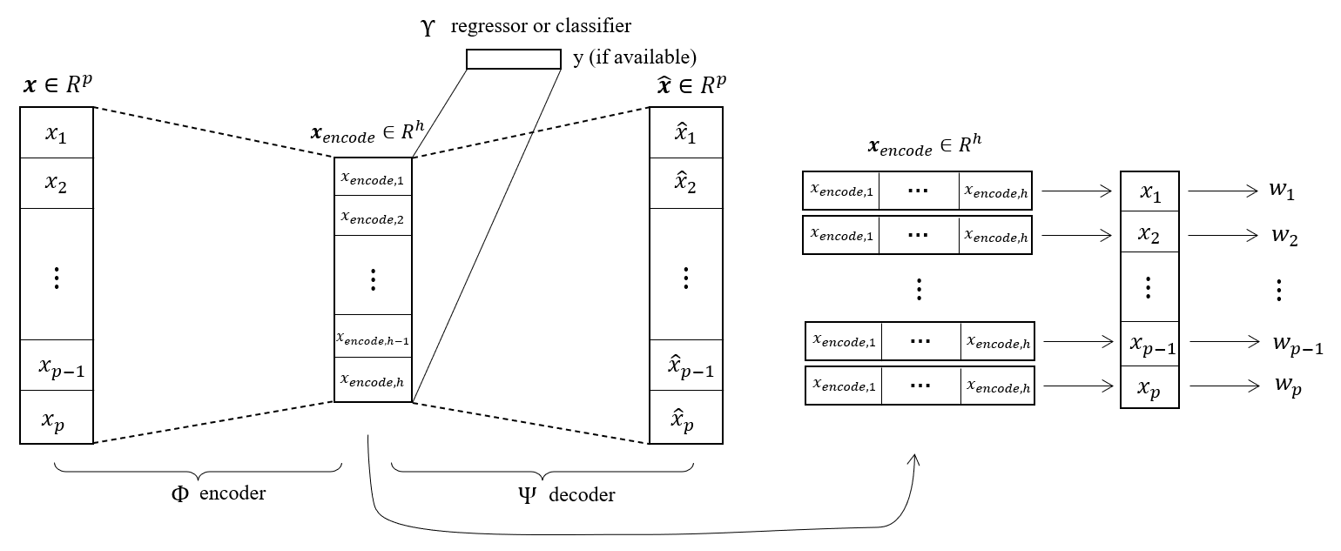}
  \caption{The architecture DeepFS consists of an autoencoder based dimension reduction method for feature extraction (on the left) and a multivariate feature screening (on the right). $\boldsymbol{x}_{\textrm{encode}} \in \mathbb{R}^h$ is the low-dimensional  representation of the original input that captures most of the information of the data in $\mathbb{R}^p$, $h \ll p$, and is used to compute the strength of dependence with each feature.}  \label{figure:structure}
\end{figure}

Figure \ref{figure:structure} outlines the architecture of the feature extraction step and the feature screening step. In the feature extraction step, we extract a good representation of data using deep learning based algorithms. In the feature screening step, we compute the importance score of each original input features by means of multivariate rank distance correlation. Furthermore, we also summarize in Algorithm \ref{algo} the detailed procedure that carries out the proposed method.

\begin{algorithm}\caption{Deep Feature Screening}\label{algo}
\KwIn{input design matrix $\boldsymbol{X}\in\mathbb{R}^{n\times p}$, response vector $\boldsymbol{y} = (y_1, \ldots, y_n)^{\top}$ (if available), encoder network $\Phi$, decoder network $\Psi$, regressor or classifier $\Upsilon$ (if available), learning rate $\eta$, tuning parameter $\lambda$, number of epochs for the first step $E$}
\textbf{Step 1: Dimension Reduction and Feature Extraction}\\
Initialize $\Theta$.\\
\For {$e \in \{1,...,E\}$}{
$\widehat{\boldsymbol{x}}_{\textrm{encode}} = \Phi(\textrm{x})$\\
$\widehat{\boldsymbol{y}} = \Upsilon(\widehat{\boldsymbol{x}}_{\textrm{encode}})$ (if response is available)\\
$\widehat{\boldsymbol{x}} = \Psi(\widehat{\boldsymbol{x}}_{\textrm{encode}})$\\
\If {response is not available:}  {$\mathcal{L} = ||\widehat{\boldsymbol{x}}-\boldsymbol{x}||_2^2$}
\If {response is continuous:}  {$\mathcal{L}$ is defined in (\ref{Eq Continuous loss})}
\If {response is categorical:}  {$\mathcal{L}$ is defined in (\ref{Eq Coategorical loss})}
Compute gradient of the loss $\mathcal{L}$ w.r.t. $\Theta$ using backpropagation $\nabla_\Theta\mathcal{L}$ and update $\Theta \leftarrow \Theta - \eta\nabla_\Theta\mathcal{L}$.
}
Finish Step 1 and obtain $\boldsymbol{x}_{\textrm{encode}} = \frac{\Phi(\boldsymbol{x}) - \min\Phi(\boldsymbol{x})}{\max\Phi(\boldsymbol{x}) - \min \Phi(\boldsymbol{x})} \in \mathbb{R}^h$. \\
\textbf{Step 2: Feature Screening}\\
Generate a Sobol sequence $\{\mathbf{c}_1,...,\mathbf{c}_n\}\subset[0, 1]^h$\\
\For {$i \in \{1,...,p\}$}{
Compute rank distance correlation between the $i$-th feature $X_i$ and $\boldsymbol{x}_{\textrm{encode}}$ $\omega_i = \textrm{RdCorr}(X_i, \boldsymbol{x}_{\textrm{encode}})$ using (\ref{assign}), (\ref{rank}), and (\ref{RdCorr}).
}
\KwRet feature importance scores
\end{algorithm}

\section{Numerical Experiments} \label{Simulation}

In this section, we conduct extensive simulation studies to examine the performance of the proposed feature screening method on synthetic datasets.  To this end, we design four simulation experiments that correspond to four different scenarios: categorical response, continuous response, unknown structure, and unsupervised learning.  Each simulation consists of 500 independent replications.  Similar to \cite{Fan_1}, we choose the top $k = [n/\log(n)]$ features in all simulation designs, where $[a]$ denotes the integer part of $a$.

We also compare our method with some state-of-the-art  methods.  They are (1) the classical feature screening method, \emph{iterative sure independence screening} (ISIS) \cite{Fan_1}, (2) two Lasso-based (or $\ell_1$-regularized) algorithms, \emph{PLasso} \cite{pLasso} and \emph{LassoNet} \cite{lassonet}, (3) the reparametrization trick-based method,  \emph{concrete autoencoder} (CAE) \cite{CAE}, and (4) two two-step methods, \emph{feature selection network} (FsNet) \cite{FsNet} and \emph{teacher-student feature selection network} (TSFS) \cite{teacher_student}.  Among the six methods, LassoNet, CAE, FsNet, and TSFS are deep learning based.

In each simulation, we let $\mathcal{J}$ denote the set of true active variables and $\widehat{\mathcal{J}}_i$  the set of selected variables in the $i$-th replication, $i=1, 2, \ldots, 500$. The following metric is used to evaluate the performance of each method:
\begin{equation}\label{eqn.metric}
\varrho = \frac{1}{500}\sum_{i=1}^{500}\frac{|\widehat{\mathcal{J}}_i \cap \mathcal{J}|}{|\mathcal{J}|},
\end{equation}
which measures the proportion  of active variables selected out of the total amount of true active variables. Such metric has been widely used in feature selection literature (see,  for instance, \cite{Fan_1} and \cite{pLasso}).

To validate the comparison, we use the Wilcoxon method to test whether the difference between DeepFS and each of the other methods is statistically significant. As such, the following hypotheses are considered:
\begin{equation}\label{wilcoxon}
H_0: \varrho_{\textrm{DeepFS}}  = \varrho_{k} ~ \textrm{ versus } ~ H_a: \varrho_{\textrm{DeepFS}}  > \varrho_{k},
\end{equation}
where $\{\varrho_{1}, \varrho_2, \ldots, \varrho_6\} $ are the $\varrho$s associated with ISIS, PLasso, CAE, FsNet, LassoNet, and TSFS, respectively.  Denote by $p_k$ the p-value of the Wilcoxon test for a comparison between DeepFS and method $k\in \{1, 2, \ldots, 6\}$, and let
\begin{equation}\label{wilcoxon1}
p_{max} = \max_{k=1, 2, \ldots, 6}p_k.
\end{equation}
To test the statistical significance, we aim to show $p_{max} $ is less than $1\%$.

There are many hyper-parameters in our approach: the number of layers, the number of neurons in each layer, dropout probability, learning rate, the tuning parameter $\lambda$ in the loss function, and the dimension of the latent space $\mathcal{F}$. These hyper-parameters play an important role in our method. It can be difficult to assign values to the hyper-parameters without expert knowledge on the domain. In practice, it is common to use grid search, random search \cite{RandomSearch}, Bayesian optimization \cite{Bayesian_Opt}, among others, to obtain optimal values of the hyper-parameters. In this paper we search for the desirable values of the hyper-parameters by minimizing the loss over the predefined parameter candidates using a validation set. As CAE is unsupervised, we follow the instruction in  \cite{CAE} to adapt it to the supervised setting. For each simulation, we use the Adam optimizer to train the network.

\subsection{Simulation 1: Categorical Response}

In this simulation study, we consider a classification problem. Inspired by the fact that  high-dimension, low-sample-size data always occur in  genome-wide association studies (GWAS) and gene selection, we follow the simulation design of  GWAS feature selection in \cite{LassoWu} and \cite{pLasso}.
We first generate $n$ independent auxiliary random vectors $\boldsymbol{Z}_i=(Z_{i1}, \ldots, Z_{ip})^{\top}, i=1, \ldots, n$, from a multivariate Gaussian distribution $\mathcal{N}(0, \boldsymbol{\Sigma})$, where $\boldsymbol{\Sigma}$ is a $p \times p$ covariance matrix capturing the correlation between SNPs. The design of $\boldsymbol{\Sigma}$ is the same as \cite{pLasso};  that is,
\begin{equation}
\boldsymbol{\Sigma}_{ij} = \left\{
\begin{array}{clc}
1 & \text{if } i = j ,\\
\rho & \text{if } i\neq j, i,j \leq p/20, \\
0 & \text{otherwise,}
\end{array}
\right.
\end{equation}
with $\rho \in\{0, 0.5, 0.8\}$. This design allows closer SNPs to have a stronger correlation. Next, we randomly generate $p$ minor allele frequencies (MAFs) $m_1, \ldots, m_p$ from the uniform distribution $Uniform(0.05, 0.5)$ to represent the strength of heritability. For the $i$-th observation, $i=1, \ldots ,n$, the SNPs $\boldsymbol{x}_i = (x_{i1}, \ldots, x_{ip})^{\top}$ are generated according to the following rule: for $ j=1, \ldots, p,$
\begin{equation*}
    \begin{aligned}
    x_{ij} =
            \left\{
             \begin{array}{ll}
              \epsilon  &Z_{ij} \leq c_1,\\
              \epsilon +1 & c_1 < Z_{ij} < c_2,   \\
              \epsilon +2 &Z_{ij} \geq c_2,
             \end{array}
            \right.
    \end{aligned}
\end{equation*}
where $\epsilon$ is sampled from $\mathcal{N}(0, 0.1^2)$, and $c_1$ and $c_2$ are the $(1-m_j)^2$-quantile and $(1-m_j^2)$-quantile of $\{Z_{1j}, \ldots, Z_{nj}\}$, respectively.  Once $\{\boldsymbol{x}_i, i=1, \ldots, n\}$ are generated, they are kept the same throughout the 500 replications.

Suppose that there are only $10$ (out of $p$) active SNPs associated with the response. Let $J = \{j_1, \ldots, j_{10}\}$, where $j_k, k=1,\ldots,10$, are randomly sampled from $\{1, \ldots, p\}$ without replacement; and once generated, the set $J$ is kept the same throughout the 500 replications.  The true relationship between the response and the features are nonlinear and determined through the following dichotomous phenotype model.  That is, for the $i$-th observation, $i=1, \ldots ,n$, the response $y_i\sim Binomial(1; \pi_i)$ and
\begin{align}
\log\frac{\pi_i}{1-\pi_i} &= -3 + \beta_1x_{i, j_1} + \beta_2\sin(x_{i, j_2}) + \beta_3\log(x_{i, j_3}^2+1) + \beta_4x_{i, j_4}^2 \nonumber\\
 &\quad + \beta_5\mathbbm{1}(x_{i, j_5}<1) + \beta_6\max(x_{i, j_6}, 1) + \beta_7x_{i, j_7}\mathbbm{1}(x_{i, j_7}<0) \nonumber\\
 &\quad + \beta_8\sqrt{|x_{i, j_8}|} + \beta_9\cos(x_{i, j_9}) + \beta_{10}\tanh(x_{i, j_{10}}), \label{eqn.sim1}
\end{align}
with $\beta_j$, $j=1, \ldots, 10$, sampled from $Uniform(1, 2)$; that is, once  $\beta_j$, $j=1, \ldots, 10$, are generated, they are kept fixed throughout the 500 replications.  To examine the effects of  sample size and dimension on the accuracy of the feature selection procedure, we consider $n\in \{200, 500, 1000\}$ and $p\in \{1000, 3000\}$.

For each replication and each combination of $\rho, n$, and $p$, we apply the seven methods to the simulated data and perform feature selection. Table \ref{Tab:Sim1_classification} reports  the proportion of  features selected by the various methods out of 10 true active features averaged over 500 replications (i.e., the statistic $\varrho$ defined in \eqref{eqn.metric}) and its standard error (displayed in parentheses) using different combinations of $\rho, n$, and $p$.   When $\rho=0$, $n=200$, and $p = 1000$,  the proposed method, DeepFS,  can select 66\% of the true active  features, FsNet comes next, while ISIS selects 46\% of them. As expected, when the sample size increases, the performance of all the seven methods improves; in particular, when $n=1000$, DeepFS and LassoNet outperform the others, with a percentage of 84\% and 81\%, respectively.  A large dimension $p$ and/or a strong dependence $\rho$  among the SNPs have an adverse impact on performance of feature selection methods. The p-values are all less than $1\%$ confirms that DeepFS always outperforms the other methods in all aspects.

\begin{table}[!t]
\centering
\caption{Results of Simulation 1: the averaged $\varrho$ over 500 replicates (with its standard error in parentheses) of various methods using different combinations of $\rho, n$, and $p$.  (+) means that $p_{\textrm{max}} < 0.01$, where $p_{\textrm{max}}$ is the maximum $p$-values for the six tests defined in \eqref{wilcoxon1}.}
\label{Tab:Sim1_classification}
{\begin{tabular}{@{}ccccccccc@{}}
\hline
$\rho$ &($n, p$) & ISIS   & PLasso & CAE   & FsNet & LassoNet & TSFS & DeepFS \\
0 &(200, 1,000) & 0.46   & 0.55   & 0.57  & 0.64  & 0.57     & 0.50 & {\bf 0.66} (+) \\
  &           & (0.05) & (0.05) & (0.06)& (0.05)& (0.05)   &(0.06)& (0.04)  \\
0 &(500, 1,000) & 0.52   & 0.66   & 0.66  & 0.70  & 0.67     & 0.55 & {\bf0.75} (+)\\
  &            & (0.05) & (0.04) & (0.05)& (0.05)& (0.05)   &(0.06)& (0.05) \\
0 &(1,000, 1,000)& 0.57   & 0.75   & 0.73  & 0.76  & 0.81     & 0.63 & {\bf 0.84} (+) \\
  &            & (0.04) & (0.04) & (0.05)& (0.04)& (0.05)   &(0.05)& (0.04) \\
\hline
\hline
$\rho$ &($n, p$) & ISIS   & PLasso & CAE   & FsNet & LassoNet &TSTS  & DeepFS\\
0 &(200, 3,000) & 0.39   & 0.42   & 0.55  & 0.54  & 0.56     & 0.38 & {\bf0.63} (+)\\
  &            & (0.06) & (0.06) & (0.05)& (0.04)& (0.06)   &(0.06)& (0.05)  \\
0 &(500, 3,000) & 0.42   & 0.59   & 0.63  & 0.65  & 0.69     & 0.52 & {\bf0.70} (+)\\
  &            & (0.06) & (0.05) & (0.05)& (0.06)& (0.05)   &(0.06)& (0.05)  \\
0 &(1,000, 3,000)& 0.46   & 0.62   & 0.68  & 0.69  & 0.71     & 0.54 & {\bf0.77} (+)\\
  &            & (0.06) & (0.05) & (0.06)& (0.05)& (0.05)   &(0.06)& (0.04)  \\
\hline
\hline
$\rho$ &($n, p$)   & ISIS   & PLasso & CAE   & FsNet & LassoNet &TSTS  & DeepFS \\
0.5 &(200, 1,000) & 0.46   & 0.46   & 0.55  & 0.59  & 0.53     & 0.47 & {\bf0.68} (+)\\
  &              & (0.06) & (0.06) & (0.05)& (0.05)& (0.04)   &(0.05)& (0.05)   \\
0.5 &(500, 1,000) & 0.49   & 0.54   & 0.66  & 0.65  & 0.68     & 0.56 & {\bf0.72}  (+)\\
  &              & (0.05) & (0.06) & (0.06)& (0.05)& (0.05)   &(0.05)& (0.04)  \\
0.5 &(1,000, 1,000)& 0.52  & 0.66   & 0.72  & 0.71  & 0.77     & 0.63 & {\bf0.81}  (+)\\
  &              & (0.06) & (0.05) & (0.05)& (0.04)& (0.05)   &(0.06)& (0.05)  \\
\hline
\hline
$\rho$ &($n, p$)   & ISIS   & PLasso & CAE   & FsNet & LassoNet &TSTS  & DeepFS \\
0.5 &(200, 3,000) & 0.36   & 0.42   & 0.46  & 0.51  & 0.46     & 0.40 & {\bf0.56} (+)\\
    &            & (0.05) & (0.04) & (0.05)& (0.05)& (0.06)   &(0.04)& (0.05)  \\
0.5 &(500, 3,000) & 0.40   & 0.49   & 0.53  & 0.60  & 0.60     & 0.49 & {\bf0.71} (+)\\
    &            & (0.07) & (0.05) & (0.06)& (0.06)& (0.07)   &(0.05)& (0.06)   \\
0.5 &(1,000, 3,000)& 0.46  & 0.57   & 0.61  & 0.67  & 0.65     & 0.53 & {\bf0.75} (+)\\
    &            & (0.05) & (0.06) & (0.05)& (0.07)& (0.06)   &(0.06)& (0.05)  \\
\hline
\hline
$\rho$ &($n, p$)  & ISIS   & PLasso & CAE   & FsNet & LassoNet &TSTS  & DeepFS \\
0.8 &(200, 1,000) & 0.34   & 0.41   & 0.54  & 0.53  & 0.54     & 0.42 & {\bf0.59} (+)\\
    &            & (0.06) & (0.05) & (0.05)& (0.05)& (0.06)   &(0.04)& (0.05)  \\
0.8 &(500, 1,000) & 0.42   & 0.51   & 0.56  & 0.68  & 0.58     & 0.45 & {\bf0.66} (+)\\
    &            & (0.07) & (0.06) & (0.06)& (0.06)& (0.07)   &(0.06)& (0.06)  \\
0.8 &(1,000, 1,000)& 0.45  & 0.59   & 0.62  & 0.70  & 0.72     & 0.55 & {\bf0.78} (+)\\
    &            & (0.06) & (0.06) & (0.06)& (0.07)& (0.08)   &(0.06)& (0.07)  \\
\hline
\hline
$\rho$ &($n, p$)   & ISIS   & PLasso & CAE   & FsNet & LassoNet &TSTS  & DeepFS \\
0.8 &(200, 3,000) & 0.27   & 0.34   & 0.47  & 0.46  & 0.46     & 0.31 & {\bf0.52} (+)\\
    &            & (0.05) & (0.04) & (0.04)& (0.04)& (0.05)   &(0.03)& (0.04) \\
0.8 &(500, 3,000) & 0.35   & 0.44   & 0.52  & 0.54  & 0.55     & 0.40 & {\bf0.63} (+)\\
    &            & (0.05) & (0.04) & (0.04)& (0.05)& (0.05)   &(0.04)& (0.06) \\
0.8 &(1,000, 3,000)& 0.39   & 0.47  & 0.56  & 0.59  & 0.62     & 0.43 &{\bf 0.68} (+)\\
    &            & (0.06) & (0.05) & (0.04)& (0.05)& (0.05)   &(0.05)& (0.07)  \\
\hline
\end{tabular}}{}
\end{table}

\subsection{Simulation 2: Continuous Response}
In this experiment, we investigate the validity of our method when applied to continuous response. Because FsNet is designed for categorical response, to provide a fair comparison, we change the classifier to a regressor in the loss function of FsNet.  The data are generated in the following way.  We draw $n$  vectors $\boldsymbol{x}_i = (x_{i1}, \ldots, x_{ip})^{\top} $, $i=1, 2, \ldots, n$, independently from a multivariate Cauchy distribution with mean zero and covariance matrix $\boldsymbol{\Sigma}_{p\times p} = (\sigma_{ij})_{p\times p}$, with $\sigma_{ij} = \rho^{|i-j|}$. The response variable $y_i$ is generated from the following model:
\begin{align}\label{eqn.sim2}
y_i = \beta_1x_{i, j_1} + \beta_2x_{i, j_2}^2 + \beta_3x_{i, j_3}x_{i, j_4} + \beta_3x_{i, j_5}x_{i, j_6} + \beta_4\mathbbm{1}(x_{i, j_6}<0) + \epsilon_i,
\end{align}
where $\beta_1, \beta_2, \beta_3, \beta_4 \sim Uniform(1, 2)$, $j_k, k=1,\ldots,6$, are sampled from $\{1, \ldots, p\}$ without replacement, and $\epsilon_i \overset{iid}\sim \mathcal{N}(0, 1)$ is the random error. Once generated, $\beta_j, j=1, \ldots, 4$, $j_k, k=1,\ldots,6$, and $\boldsymbol{x}_i, i=1, \ldots, n$, are all kept the same throughout the 500 replications. So in this experiment, the response variable is related to six active features via Model \eqref{eqn.sim2}.

 Same as Simulation 1, we take $\rho \in\{0, 0.5, 0.8\}$, $n\in \{200, 500, 1000\}$, and $p\in \{1000, 3000\}$.  We apply the seven methods to each of the 500 datasets generated from Model \eqref{eqn.sim2}, and the feature selection results are summarized in Table \ref{Tab:Sim2_regression}, in which we report the percentage of the selected true features out of six and its standard error, and whether the maximum $p$-value for the six tests defined in \eqref{wilcoxon} is less than 1\%.  The highest percentage among all the methods is highlighted in boldface for each combination of $\rho$, $n$, and $p$.  Consistent with the observations in Table \ref{Tab:Sim1_classification}, DeepFS is the best among all the  methods, no matter the sample size, the dimension of the features, or whether the features are dependent. The other six methods perform very similar to each other.

\begin{table}[!t]
\centering
\caption{Results of Simulation 2: the averaged $\varrho$ over 500 replicates (with its standard error in parentheses) of various methods using different combinations of $\rho, n$, and $p$.  (+) means that $p_{\textrm{max}} < 0.01$, where $p_{\textrm{max}}$ is the maximum $p$-values for the six tests defined in \eqref{wilcoxon1}. }
\label{Tab:Sim2_regression}
{\begin{tabular}{@{}ccccccccc@{}}
\hline
$\rho$ &($n, p$) & ISIS   & PLasso & CAE   & FsNet & LassoNet & TSFS & DeepFS \\
0 &(200, 1,000) & 0.45   & 0.43   & 0.36  & 0.44  & 0.42     & 0.42 & {\bf0.59} (+)   \\
  &            & (0.05) & (0.04) & (0.04)& (0.05)& (0.05)   &(0.04)& (0.04) \\
0 &(500, 1,000) & 0.48   & 0.48   & 0.51  & 0.52  & 0.57     & 0.48 & {\bf0.65} (+)\\
  &            & (0.06) & (0.05) & (0.04)& (0.05)& (0.03)   &(0.04)& (0.04) \\
0 &(1,000, 1,000)& 0.51   & 0.54   & 0.56  & 0.60  & 0.62     & 0.50 & {\bf0.71} (+)\\
  &            & (0.05) & (0.04) & (0.04)& (0.04)& (0.03)   &(0.05)& (0.05) \\
\hline
\hline
$\rho$ &($n, p$) & ISIS   & PLasso & CAE   & FsNet & LassoNet & TSFS & DeepFS \\
0 &(200, 3,000) & 0.35   & 0.34   & 0.34  & 0.35  & 0.37     & 0.36 & {\bf0.53} (+)\\
  &            & (0.05) & (0.04) & (0.05)& (0.05)& (0.05)   &(0.04)& (0.05) \\
0 &(500, 3,000) & 0.38   & 0.39   & 0.41  & 0.44  & 0.47     & 0.41 & {\bf0.55} (+)\\
  &            & (0.05) & (0.05) & (0.05)& (0.06)& (0.05)   &(0.05)& (0.04) \\
0 &(1,000, 3,000)& 0.43   & 0.46   & 0.44  & 0.53  & 0.56     & 0.47 & {\bf0.64} (+) \\
  &            & (0.06) & (0.06) & (0.05)& (0.05)& (0.04)   &(0.06)& (0.05) \\
\hline
\hline
$\rho$   &($n, p$) & ISIS   & PLasso & CAE   & FsNet & LassoNet & TSFS & DeepFS\\
0.5 &(200, 1,000) & 0.37   & 0.35   & 0.38  & 0.36  & 0.38     & 0.33 & {\bf0.50} (+)\\
    &            & (0.05) & (0.05) & (0.06)& (0.05)& (0.05)   &(0.04)& (0.04) \\
0.5 &(500, 1,000) & 0.41   & 0.44   & 0.42  & 0.40  & 0.42     & 0.37 & {\bf0.57} (+)\\
    &            & (0.06) & (0.05) & (0.05)& (0.07)& (0.06)   &(0.06)& (0.05) \\
0.5 &(1,000, 1,000)& 0.44   & 0.47   & 0.48  & 0.45  & 0.54     & 0.53 & {\bf0.65} (+)\\
    &            & (0.06) & (0.05) & (0.06)& (0.07)& (0.05)   &(0.08)& (0.05) \\
\hline
\hline
$\rho$   &($n, p$) & ISIS   & PLasso & CAE   & FsNet & LassoNet & TSFS & DeepFS\\
0.5 &(200, 3,000) & 0.33   & 0.32   & 0.29  & 0.34  & 0.30     & 0.27 & {\bf0.43} (+)\\
    &            & (0.04) & (0.04) & (0.05)& (0.04)& (0.05)   &(0.04)& (0.05) \\
0.5 &(500, 3,000) & 0.36   & 0.36   & 0.35  & 0.43  & 0.35     & 0.36 & {\bf0.47} (+)\\
    &            & (0.05) & (0.05) & (0.05)& (0.05)& (0.06)   &(0.06)& (0.05) \\
0.5 &(1,000, 3,000)& 0.40  & 0.43   & 0.41  & 0.45  & 0.42     & 0.43 & {\bf0.51} (+)\\
    &            & (0.05) & (0.06) & (0.05)& (0.04)& (0.04)   &(0.05)& (0.04) \\
\hline
\hline
$\rho$   &($n, p$)   & ISIS   & PLasso & CAE   & FsNet & LassoNet & TSFS & DeepFS\\
0.8 &(200, 1,000)  & 0.32  & 0.32   & 0.34  & 0.33  & 0.32     & 0.33 & {\bf0.46} (+)\\
    &              & (0.03) & (0.04) & (0.03)& (0.05)& (0.04)   &(0.04)& (0.04) \\
0.8 &(500, 1,000)  & 0.36   & 0.38   & 0.36  & 0.37  & 0.36    & 0.32 & {\bf0.53} (+)\\
    &              & (0.04) & (0.05) & (0.04)& (0.04)& (0.05)   &(0.05)& (0.05) \\
0.8 &(1,000, 1,000)& 0.40  & 0.43   & 0.42  & 0.43  & 0.45     & 0.42 & {\bf0.59} (+)\\
    &              & (0.04) & (0.05) & (0.04)& (0.05)& (0.05)   &(0.05)& (0.05) \\
\hline
\hline
$\rho$ &($n, p$)     & ISIS   & PLasso & CAE   & FsNet & LassoNet & TSFS & DeepFS\\
0.8 &(200, 3,000)  & 0.26   & 0.30   & 0.26  & 0.32  & 0.29     & 0.28 & {\bf0.37} (+)\\
    &              & (0.03) & (0.03) & (0.03)& (0.04)& (0.04)   &(0.04)& (0.03) \\
0.8 &(500, 3,000)  & 0.31   & 0.34   & 0.32  & 0.39  & 0.38     & 0.32 & {\bf0.45} (+)\\
    &              & (0.03) & (0.04) & (0.04)& (0.05)& (0.03)   &(0.04)& (0.04) \\
0.8 &(1,000, 3,000)& 0.36   & 0.37   & 0.37  & 0.43  & 0.42     & 0.39 & {\bf 0.49} (+)\\
    &              & (0.04) & (0.04) & (0.05)& (0.04)& (0.05)   &(0.04)& (0.03) \\
\hline
\end{tabular}}{}
\end{table}

\subsection{Simulation 3: Unknown Structure}
In this simulation, we consider the \emph{Breast Cancer Coimbra Data Set} from UCI Machine Learning Repository \url{https://archive.ics.uci.edu/ml/datasets/\\Breast+Cancer+Coimbra}.
The dataset consists of nine quantitative predictors (or features) and a binary response that indicates the presence or absence of breast cancer, with a total of  $n=116$ observations.

Note that the true relationship between  the response and the predictors is unknown in this simulation, and that it is reasonable to assume that these nine  predictors have an association with the response.  The following procedure is adopted in order to effectively evaluate our method. We add $p-9$ more irrelevant features that are independently  sampled from standard normal distribution and are independent of the response. In other words, among all the $p$ features, there are only nine active features associated with the response, while all the others are irrelevant. The goal of this simulation is to check whether our method can correctly select these nine features from a high-dimensional feature space.

We report in Table \ref{Tab:Sim3_unkonwn_structure}  the statistic $\varrho$  based on 500 replications and its standard error, and whether the maximum $p$-value for the six tests defined in \eqref{wilcoxon} is less than 1\% for $p = 1000,  5000$, and $10000$. When $p=1000$, our method can select as high as 62\% of the true active predictors and TSFS comes next, whereas ISIS and PLasso select around 20\% of the true predictors; and the difference between DeepFS and each of the other methods is statistically significant. When the feature dimension $p$ increases, the selection precision of various methods drops, but a similar pattern emerges: the deep learning based methods (i.e., CAE, FsNet, LassoNet, TSFS, and DeepFS) always perform much better than the classical feature selection methods (i.e., ISIS and PLasso), and the proposed method is more capable of selecting the right active predictors than the others, regardless of the dimensionality.


\begin{table}[!t]
\centering
\caption{Results of Simulation 3: the averaged $\varrho$ over 500 replicates (with its standard error in parentheses) of various methods using different $p$s.  (+) means that $p_{\textrm{max}} < 0.01$, where $p_{\textrm{max}}$ is the maximum $p$-values for the six tests defined in \eqref{wilcoxon1}. }
\label{Tab:Sim3_unkonwn_structure}
{\begin{tabular}{@{}cccccccc@{}}
\hline
($n, p$)      & ISIS   & PLasso & CAE   & FsNet & LassoNet & TSFS & DeepFS \\
(116, 1,000)  & 0.20   & 0.18   & 0.26  & 0.27  & 0.31     & 0.54    & {\bf0.62} (+)\\
              & (0.02) & (0.03) & (0.03)& (0.03)& (0.04)   & (0.04)  &(0.04)\\
(116, 5,000)  & 0.18   & 0.15   & 0.24  & 0.24  & 0.29     & 0.51    &{\bf0.55}  (+)\\
              & (0.03) & (0.03) & (0.02)& (0.02)& (0.04)   & (0.03)  &(0.04)\\
(116, 10,000) & 0.13   & 0.13   & 0.21  & 0.21  & 0.25     & 0.49    &{\bf0.51} (+)\\
              & (0.02) & (0.02) & (0.02)& (0.02)& (0.03)   & (0.04)  &(0.04)\\
\hline
\end{tabular}}{}
\end{table}

\subsection{Simulation 4: Unsupervised Learning}
Our last simulation investigates the empirical performance of unsupervised learning. Because ISIS and PLasso only work for supervised learning, we omit these two methods from analysis and focus on CAE, FsNet, LassoNet, TSFS, and DeepFS.

Suppose there are two classes of data with sample sizes $n_1 = n_2 = n$, and dimension $p$. The data in the two classes are drawn from two independent multivariate Gaussian distributions: $\mathcal{N}_p(\boldsymbol{0}_p, \boldsymbol{I}_p)$ for Class 1 and $\mathcal{N}_p(\boldsymbol{\mu}, \boldsymbol{I}_p)$ for Class 2, where $\boldsymbol{0}_p$ is a $p$ dimensional vector of zeros and $\boldsymbol{I}_p$ is a $p \times p$ identity matrix.  Moreover, $\boldsymbol{\mu}$ is a $p$ dimensional vector with the $j_k$-th element equal one for $k=1, \ldots,10$ and the rest all being zeros, where $j_k$s are drawn from $\{1, \ldots, p\}$ without replacement and are kept fixed once generated. In other words, the mean vector in the second class has 10 non-zero values and their locations are the same across the 500 replications.

As in Simulations 1 and 2, we compute the selection precision statistic $\varrho$ and its standard error with $n\in \{200, 500, 1000\}$ and $p\in \{1000, 3000\}$ based on 500 replications, as well as  whether the maximum $p$-value for the four tests defined in \eqref{wilcoxon} is less than 1\% for each combination.   The results are summarized in Table \ref{Tab:Sim4_unurpervised}, where again we highlight the largest percentage among the five methods for each combination of $n$ and $p$. The numerical results indicate the superiority of our method, DeepFS, in unsupervised feature selection setting.

\begin{table}[!t]
\centering
\caption{Results of Simulation 4: the averaged $\varrho$ over 500 replicates (with its standard error in parentheses) of various methods using different combinations of $\rho, n$, and $p$. (+) means that $p_{\textrm{max}} < 0.01$, where $p_{\textrm{max}}$ is the maximum $p$-values for the six tests defined in \eqref{wilcoxon1}. }
\label{Tab:Sim4_unurpervised}
{\begin{tabular}{@{}cccccccc@{}}
\hline
($n, p$)        & ISIS   & PLasso & CAE   & FsNet & LassoNet & TSFS & DeepFS\\
(200, 1,000)    & *      & *      & 0.55  & 0.63  & 0.59     & 0.66    & \bf{0.70} (+)\\
                & *      & *      & (0.05)& (0.06)& (0.06)   & (0.05)  & (0.06)\\
(500, 1,000)    & *      & *      & 0.67  & 0.74  & 0.78     & 0.76    & \bf{0.84} (+)\\
                & *      & *      & (0.05)& (0.06)& (0.06)   & (0.05)  & (0.06)\\
(1,000, 1,000)  & *      & *      & 0.80  & 0.79  & 0.78     & 0.83    & \bf{0.88} (+)\\
                & *      & *      & (0.06)& (0.07)& (0.06)   & (0.07)  & (0.07)\\
\hline
\hline
($n, p$)        & ISIS   & PLasso & CAE   & FsNet & LassoNet & TSFS & DeepFS\\
(200, 3,000)    & *      & *      & 0.46  & 0.50  & 0.57     & 0.55    & \bf{0.62} (+)\\
                & *      & *      & (0.04)& (0.06)& (0.04)   & (0.05)  & (0.05)\\
(500, 3,000)    & *      & *      & 0.57  & 0.61  & 0.60     & 0.64    & \bf{0.71} (+)\\
                & *      & *      & (0.04)& (0.05)& (0.04)   & (0.05)  & (0.06)\\
(1,000, 3,000)  & *      & *      & 0.63  & 0.68  & 0.69     & 0.74    & \bf{0.79} (+)\\
                & *      & *      & (0.05)& (0.05)& (0.04)   & (0.06)  & (0.06)\\
\hline
\end{tabular}}{}
\end{table}

\section{Real data analysis} \label{Real_Data}
In this section, we  illustrate the usefulness of the proposed feature selection method by applying it to real-world problems. We  introduce the datasets and evaluation metrics in Section \ref{real_data_intro} and present the empirical findings in Section \ref{real_data_results}.

\subsection{Datasets and Metrics} \label{real_data_intro}
We consider four datasets that feature high dimension and low sample size and can be downloaded from \url{https://jundongl.github.io/scikit-feature}.

\begin{itemize}
\item \textbf{Colon}. The colon tissues data consist of gene expression differences between tumor and normal colon tissues measured in $p=2000$ genes, of which $20$ samples are from normal group and $42$ samples are from tumor tissues.

\item \textbf{Leukemia}. The Leukemia data are a benchmark dataset for high-dimensional data analysis which contains two classes of samples, the  acute lymphoblastic leukaemia group and the acute myeloid leukaemia group,  with observations $47$ and $25$ respectively. The dimensionality of the dataset is $p=7070$.

\item \textbf{Prostate Cancer}. The prostate cancer dataset contains genetic expression levels for $p=6033$ genes with sample size $102$, in which $50$ are normal control subjects and $52$ are prostate cancer patients.

\item \textbf{CLL\_SUB}.  The CLL\_SUB dataset has gene expressions from high density oligonucleotide arrays containing genetically and clinically distinct subgroups of B-cell chronic lymphocytic leukemia. The dataset consists of $p=11340$ attributes, 111 instances, and 3 classes.
\end{itemize}

A brief summary of the four datasets is provided in Table \ref{Tab:Real_Data_Description}. For each dataset, we run various feature selection algorithms to select the top $k$ features, with $k$ varying from 10 to $150$, and the dimension $h$ of the latent space $\mathcal{F}$ is chosen as 5  (its adequacy will be justified later via a sensitivity analysis). Following the practice in \cite{teacher_student}, \cite{lassonet}, and \cite{FsNet}, we evaluate the performance of the various methods using two criteria: classification accuracy and reconstruction error.  In order to assess classification accuracy, we use the selected features as the input for a downstream classifier: a single-layer feed-forward neural network.  The classification accuracy is measured on the test set. In addition, we also consider other classifiers, and the results are summarized in the supplementary materials. Regarding the reconstruction error,  we reconstruct the original data using the selected features and a single-layer feed-forward neural network, and define the reconstruction error as  the mean square error between the original and the reconstructed ones.

\begin{table}[]
\centering
\caption{Summary of the four datasets used in Section \ref{Real_Data}.}
\label{Tab:Real_Data_Description}
\begin{tabular}{|c|c|c|c|c|}
\hline
Dataset        & Colon & Leukemia & Prostate Cancer& CLL\_SUB  \\ \hline
Sample Size    & 62    & 72       & 102            & 111       \\ \hline
Dimensionality & 2000 & 7070    & 6033          & 11340    \\ \hline
Number of Classes&  2    &   2      &   2             &3       \\ \hline
\end{tabular}
\end{table}

\subsection{Results} \label{real_data_results}

We divide each dataset randomly into training, validation, and test sets with a 70-20-10 split. We use the training set  to learn the parameters of the autoencoders, the validation set to select the optimal hyper-parameters, and the test set to evaluate the generalization performance. We repeat the entire process 10 times, and each time we compute the statistics using the  aforementioned  criteria \citep{yamada2018ultra, CAE, lassonet}. We then average these statistics across the 10 experiments.  As the method CAE is unsupervised, we add a softmax layer to its loss function.

\begin{figure}[ht!]
  \centering
  \includegraphics[width=5in]{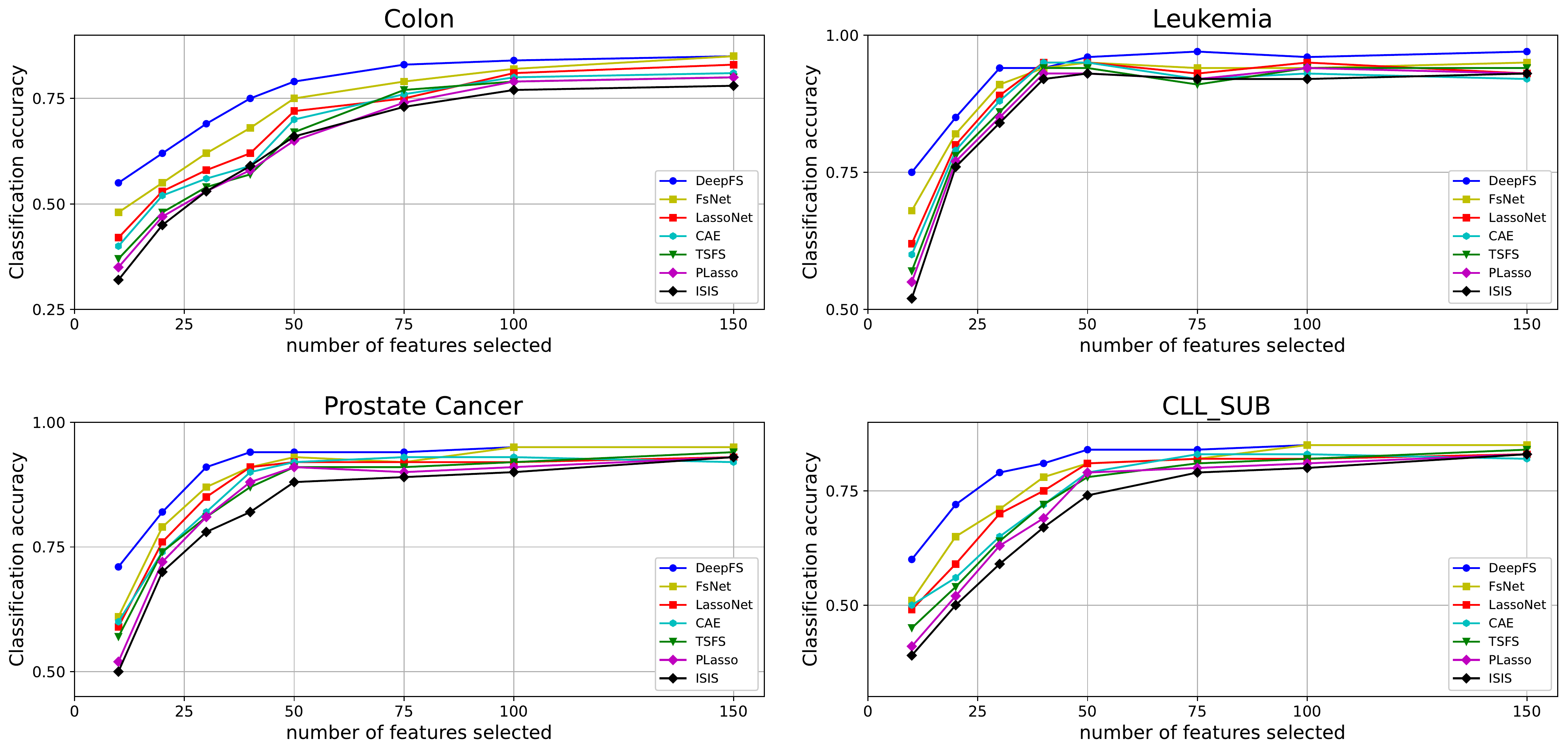}
  \caption{\it A comparison of classification accuracy among various methods. For each method, we use a single hidden layer neural network with ReLU active function for classification. All reported values are on a hold-out test set.}
  \label{figure:classification}
\end{figure}

\begin{figure}[ht!]
  \centering
  \includegraphics[width=5in]{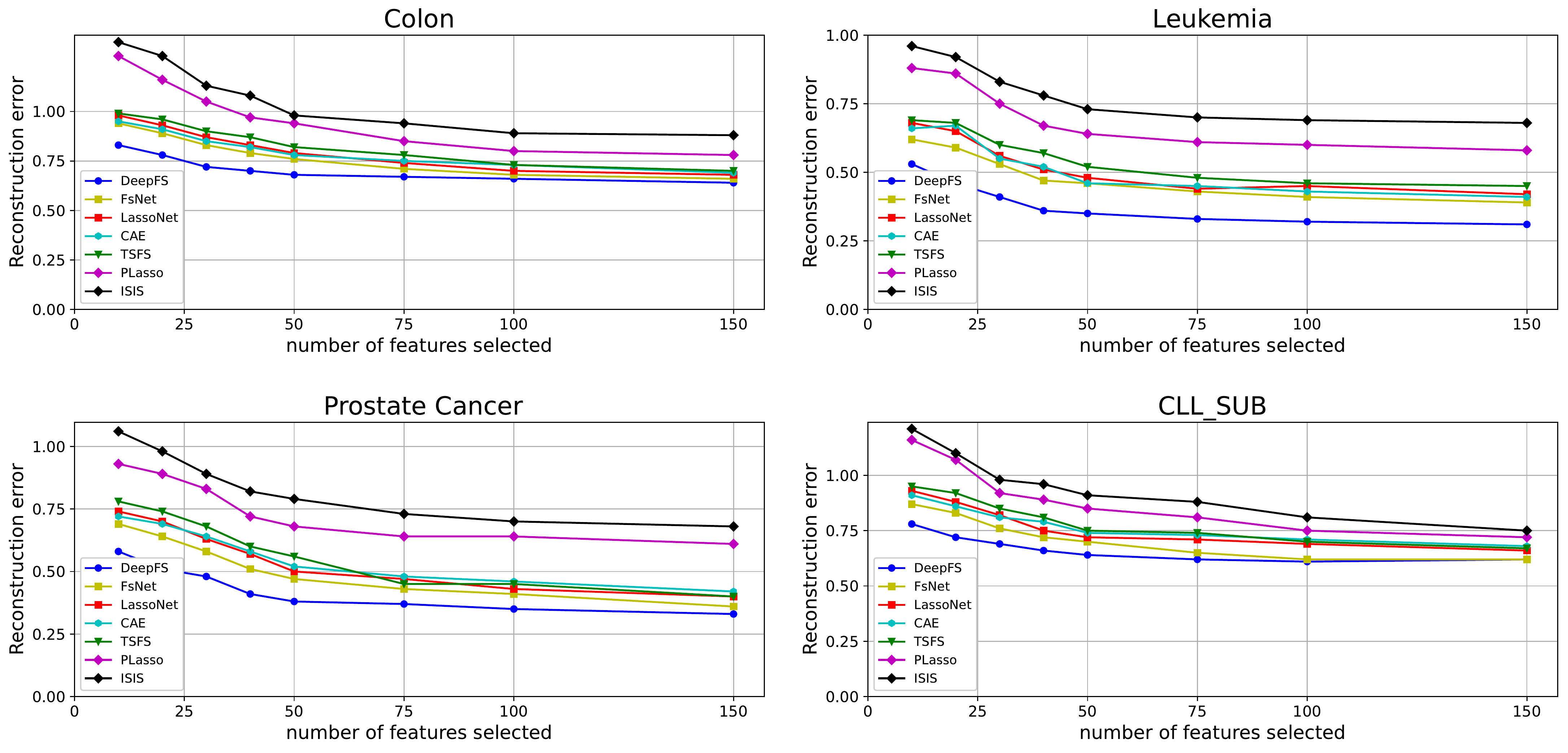}
  \caption{\it A comparison of reconstruction error among various methods. For each method, a single hidden layer neural
network with ReLU active function is employed to reconstruct the original input. All reported values are on a hold-out test set.}
  \label{figure:reconstruction}
\end{figure}

In Figure \ref{figure:classification}, we plot the classification accuracy as a function of $k$, the number of selected features. In comparison with the other methods, for all the four datasets, DeepFS has a higher classification accuracy regardless of the value of $k$ and a faster convergence rate as $k$ increases.  In particular, our method yields a much higher level of classification accuracy when the number of selected features is small. These observations suggest that the features selected by DeepFS are more informative.
Figure \ref{figure:reconstruction} shows the results of reconstruction error against the number of selected features.  It is evident that DeepFS is the best performer among all the methods across all the datasets.
Moreover,  Figures \ref{figure:classification} and \ref{figure:reconstruction} indicate that the deep learning based methods (i.e., DeepFS, FsNet, CAE,  LassoNet, and TSFS) overall have a better performance than classical feature selection algorithms (i.e., ISIS and PLasso).

\begin{figure}[ht!]
  \centering
  \includegraphics[width=5in]{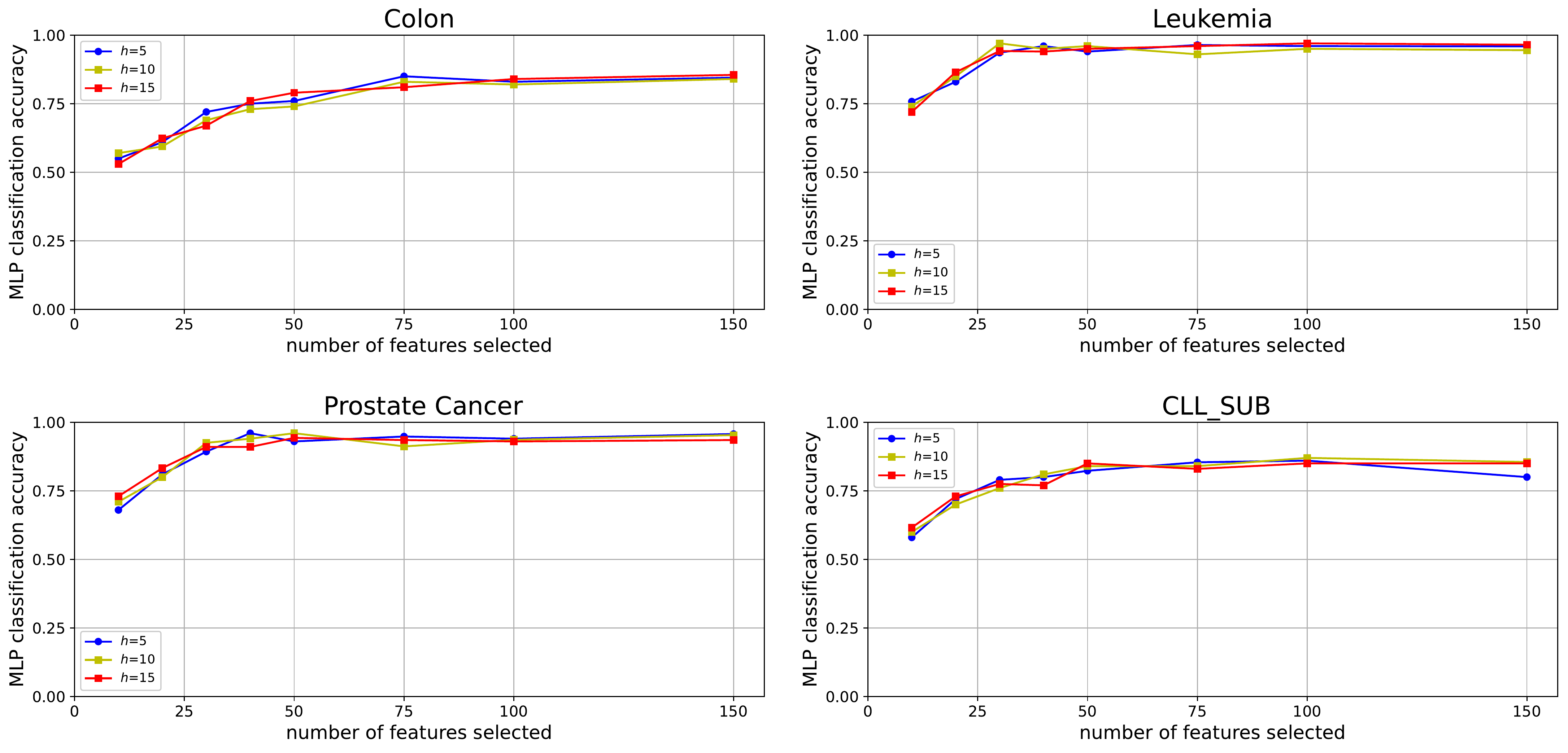}
  \caption{\it Sensitivity analysis of the dimensionality of $\boldsymbol{x}_{\textrm{encode}}$.}
  \label{figure:sensitivity}
\end{figure}

In our method, the dimensionality $h$ of the latent space $\mathcal{F}$ (i.e., the dimensionality of $\boldsymbol{x}_{\textrm{encode}}$) is a hyper-parameter controlling the relative accuracy between the feature extraction step and feature screening step. While a higher value of $h$  helps extract more information from the autoencoder, it results in a larger variance in  estimating the multivariate rank distance correlation. Hence, we perform a sensitivity analysis that investigates the effect of $h$ on the performance of DeepFS. 
Figure \ref{figure:sensitivity} depicts the classification accuracy of our method as $k$ increases from 10 to 150 with $h=5, 10$, and $15$.  The three curves associated with the different values of $h$ are entangled together, suggesting that DeepFS is insensitive to the choice of $h$.

\section{Conclusion} \label{Conclusion}

In this paper, we  proposed a new framework named DeepFS that novelly combines deep learning and feature screening for feature selection under high-dimension, low-sample-size setting. DeepFS consists of two steps: a deep neural network for feature extraction and a multivariate feature screening for feature selection. DeepFS enjoys both advantages of deep learning and feature screening. Unlike most of the existing feature screening methods, DeepFS takes into account interactions among features and the reconstruction of  original inputs, owing to the deep neural network in the first step. Moreover, DeepFS is applicable to both unsupervised and supervised settings with continuous or categorical responses.

Our numerical and empirical analyses demonstrated the superiority of DeepFS.  We are now exploring its theoretical justification and  establishing sure screening property in a follow-up paper.
In the current setup, we used an autoencoder for feature extraction, but this can be easily adapted to other methods, such as convolutional neural network, or even more sophisticated network architectures. Because the feature screening step involves calculating the rank distance correlation between  the encoded data  $\boldsymbol{x}_{\textrm{encode}}$ and each of the $p$ features, we will investigate how to speed up the computation of the second step for future research.

\bibliographystyle{apalike}
\bibliography{mainSection}
\section{Additional Results for Classification Accuracy}\label{sec.classification}

\begin{figure}[ht!]
  \centering
  \includegraphics[width=5in]{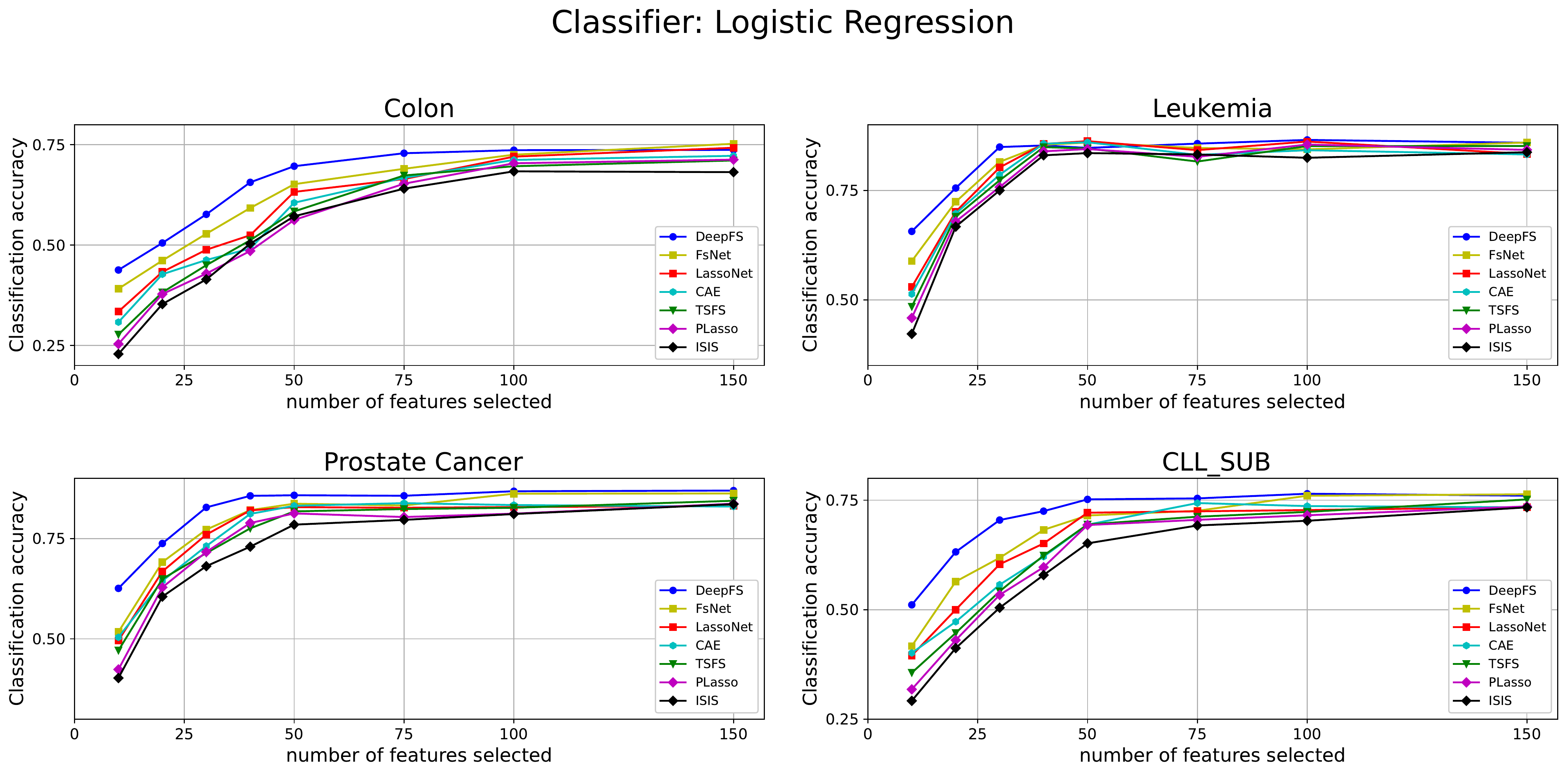}
  \caption{\it A comparison of classification accuracy among various methods. For each method, we use a logistic regression for classification. All reported values are on a hold-out test set.}
  \label{figure:classification_logistic}
\end{figure}

\begin{figure}[ht!]
  \centering
  \includegraphics[width=5in]{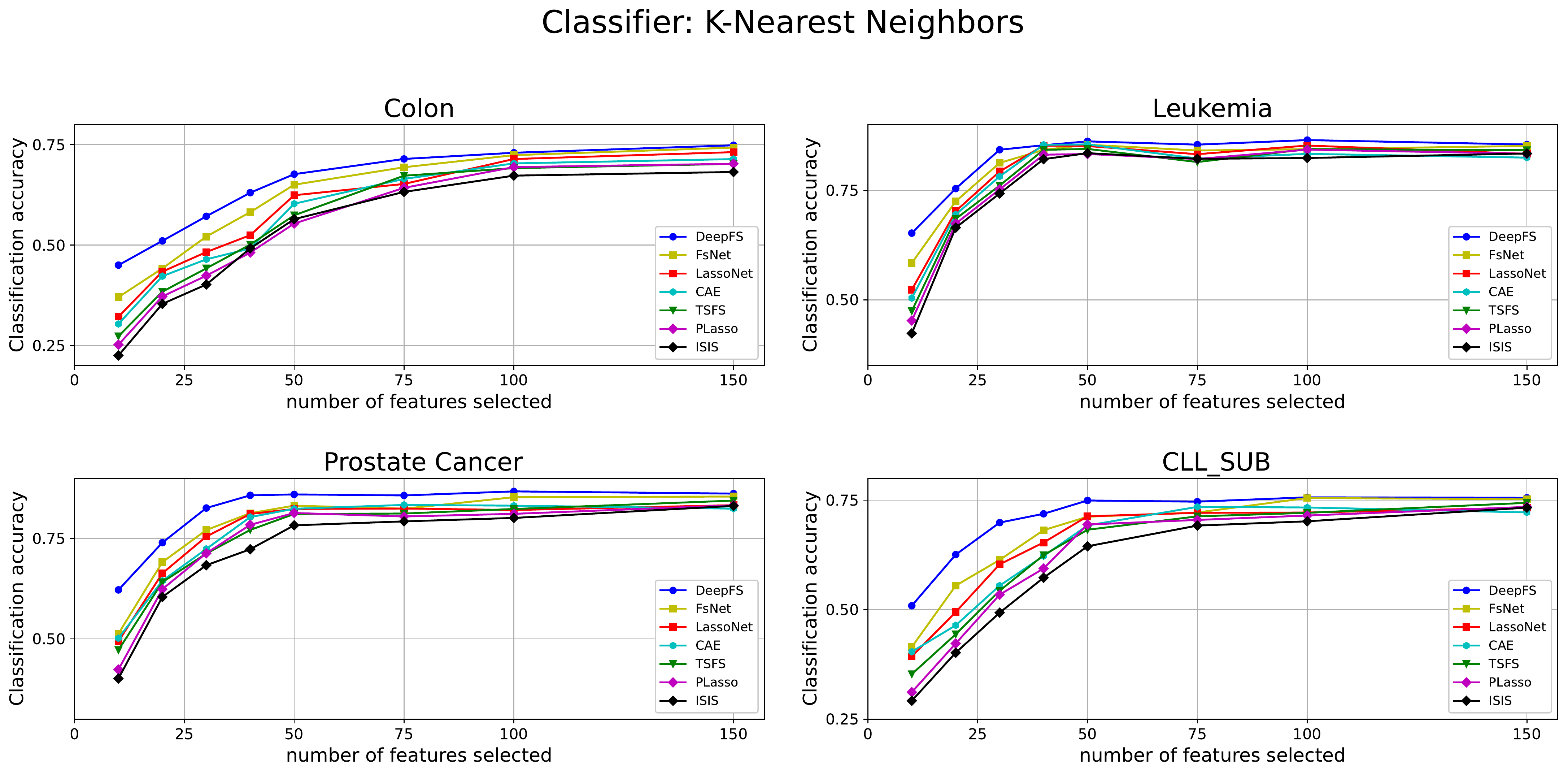}
  \caption{\it A comparison of classification accuracy among various methods. For each method, we use a K-nearest neighbors for classification. All reported values are on a hold-out test set.}
  \label{figure:classification_knn}
\end{figure}

\begin{figure}[ht!]
  \centering
  \includegraphics[width=5in]{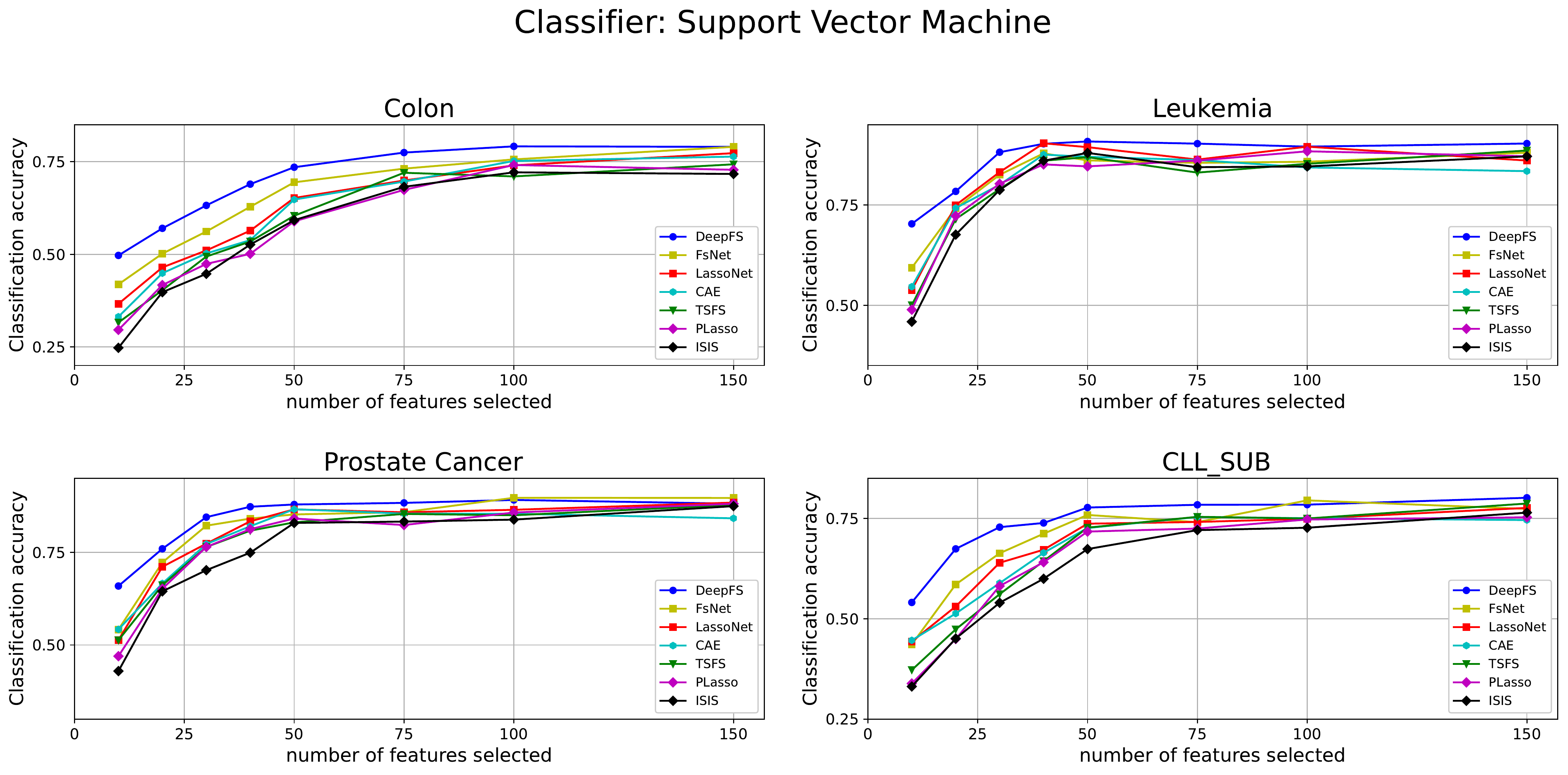}
  \caption{\it A comparison of classification accuracy among various methods. For each method, we use support vector machine for classification. All reported values are on a hold-out test set.}
  \label{figure:classification_svm}
\end{figure}

\begin{figure}[ht!]
  \centering
  \includegraphics[width=5in]{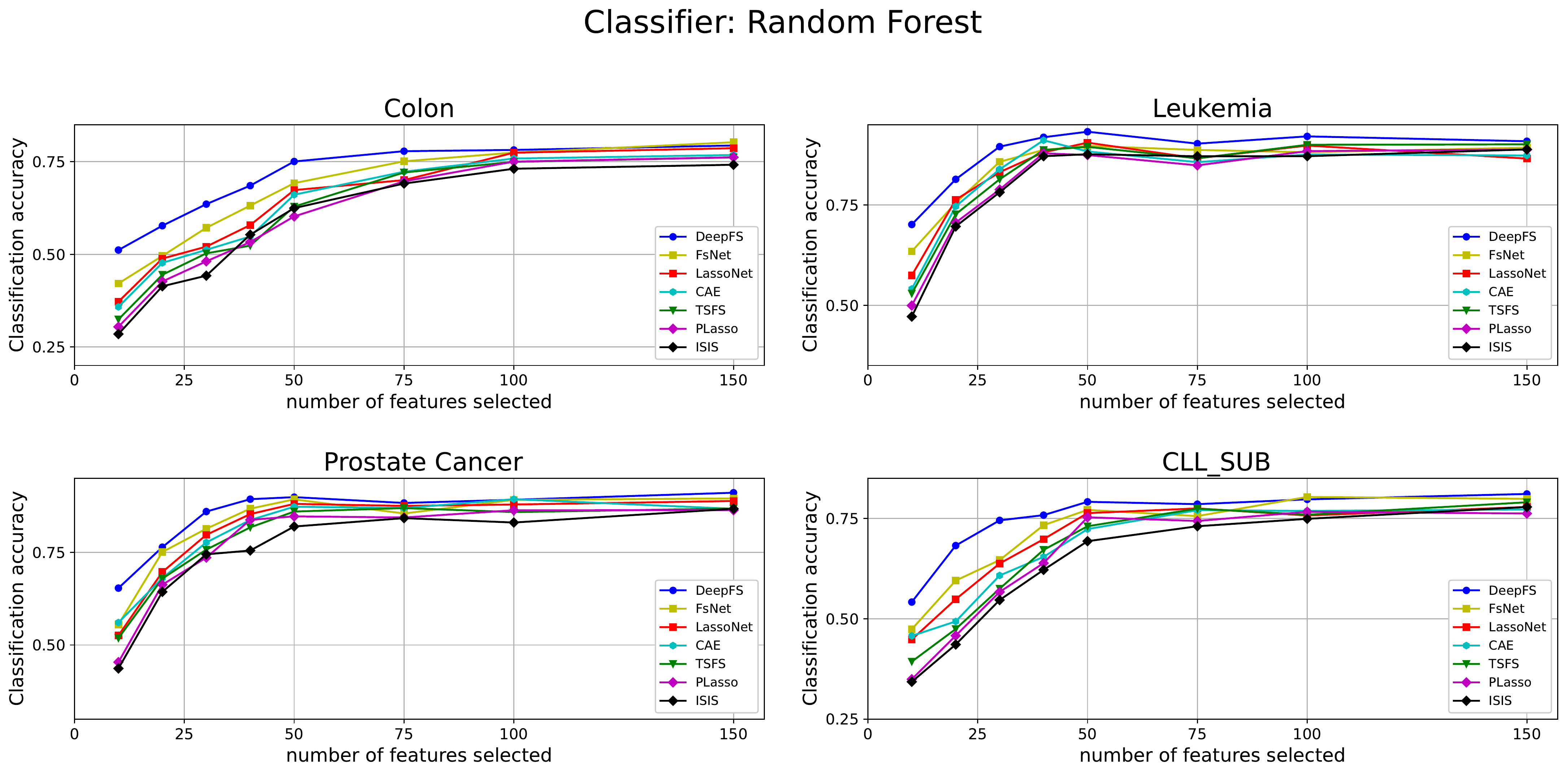}
  \caption{\it A comparison of classification accuracy among various methods. For each method, we use random forest for classification. All reported values are on a hold-out test set.}
  \label{figure:classification_RF}
\end{figure}

\begin{figure}[ht!]
  \centering
  \includegraphics[width=5in]{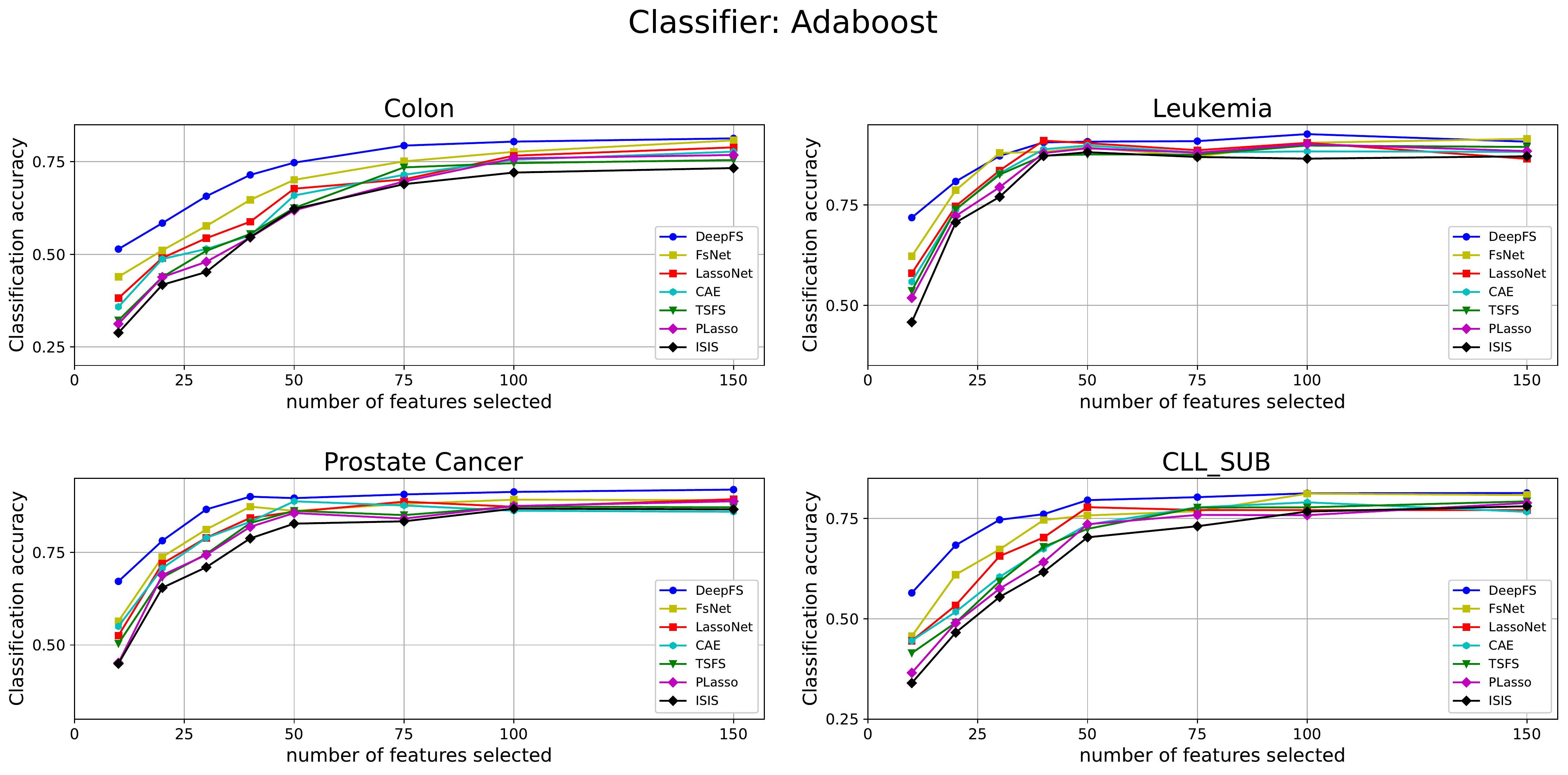}
  \caption{\it A comparison of classification accuracy among various methods. For each method, we use Adaboost for classification. All reported values are on a hold-out test set.}
  \label{figure:classification_adaboost}
\end{figure}

In the real data analysis section, we use a single-layer feed-forward neural network as the downstream classifier to assess classification accuracy. In this section, we also compare DeepFS with five other classifiers, including (multi-class) logistic regression, K-nearest neighbors, support vector machine, random forest, and AdaBoost, where the tuning parameters are selected based on cross-validation. The results for classification are summarized through Figures \ref{figure:classification_logistic}, \ref{figure:classification_knn}, \ref{figure:classification_svm}, \ref{figure:classification_RF}, and \ref{figure:classification_adaboost}. From these results, we find that our method DeepFS outperforms other methods for all classifiers.

\section{Additional Results for A Different Train-Valid-Test Split}\label{sec.split}

\begin{figure}[ht!]
  \centering
  \includegraphics[width=5in]{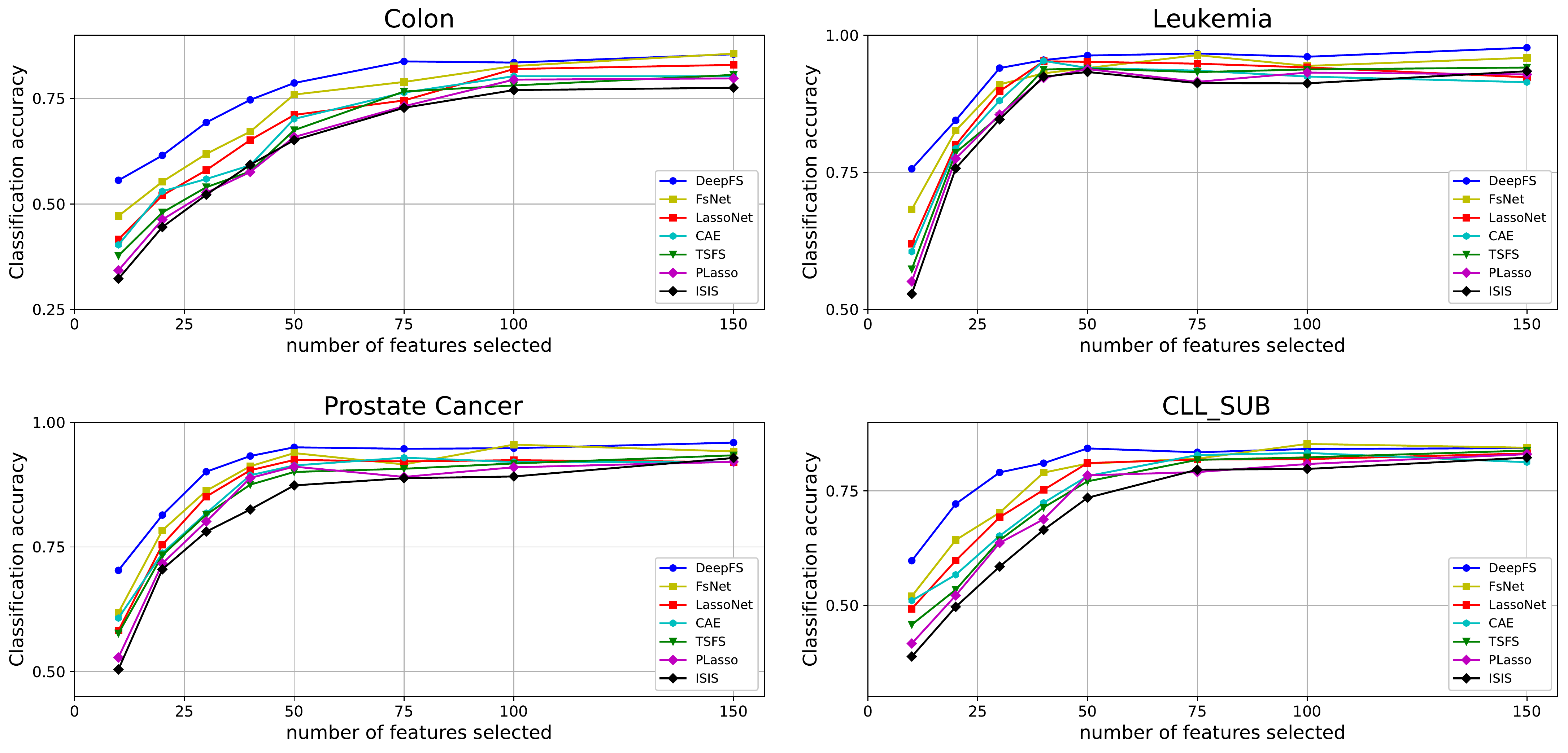}
  \caption{\it A comparison of classification accuracy among various methods. The train-valid-test split is 60-20-20. For each method, we use a single hidden layer neural network with ReLU active function for classification. All reported values are on a hold-out test set.}
  \label{figure:classification_revision}
\end{figure}

\begin{figure}[ht!]
  \centering
  \includegraphics[width=5in]{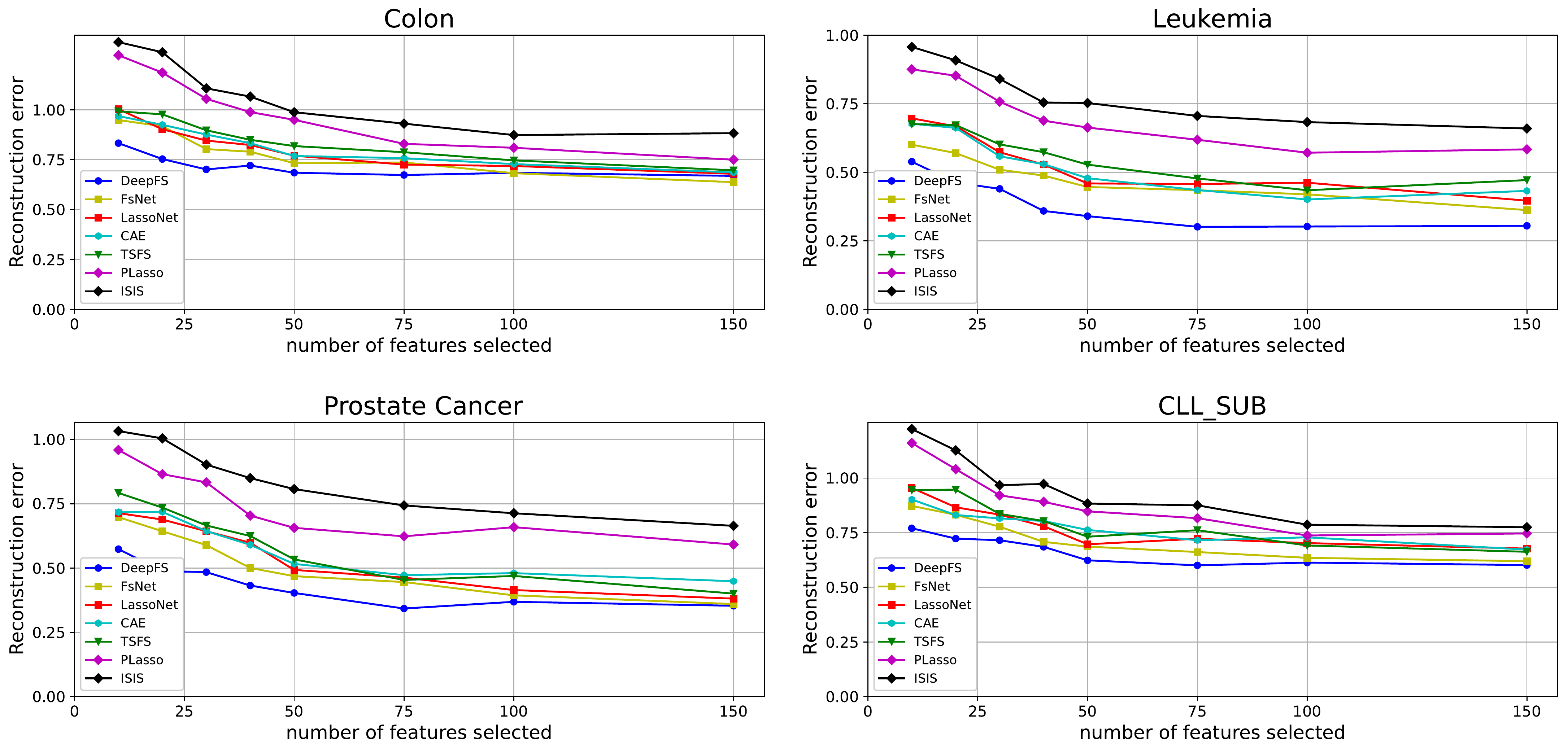}
  \caption{\it A comparison of reconstruction error among various methods. The train-valid-test split is 60-20-20. For each method, a single hidden layer neural network with ReLU active function is employed to reconstruct the original input. All reported values are on a hold-out test set.}
  \label{figure:Reconstruction_revision}
\end{figure}

In this section, we apply a different splitting scheme to the data examples.  Specifically, we divide each dataset randomly into training, validation, and test sets with a 60-20-20 split, and repeat the entire process 20 times. The results are presented in Figures \ref{figure:classification_revision} and \ref{figure:Reconstruction_revision}. They show a similar pattern to Figures 3 and 4 in the main text.

\bigskip

\end{document}